# HERMES: Heterogeneous Edge-Relational Multi-Head Embedded SSM Attention for Traffic Conflict Prediction at Signalized Intersections

**Md Monzurul Islam***
Graduate Student
Department of Civil Engineering
Texas State University
Email: monzurul@txstate.edu
ORCiD number: 0009-0007-3670-6100

**Subasish Das**
Associate Professor
Department of Civil Engineering
Texas State University
Email: subasish@txstate.edu
ORCiD number: 0000-0002-1671-2753

*Corresponding author

## ABSTRACT

Surrogate safety measures (SSMs), such as time-to-collision and deceleration rate to avoid a crash, enable proactive traffic safety analysis without relying solely on sparse and reactive crash records. However, many existing SSM-based methods evaluate pairwise interactions independently or compress multi-agent traffic scenes into flat feature vectors, limiting their ability to preserve interaction topology and evolving scene-level risk. This study addresses this limitation by formulating traffic conflict assessment as a temporal scene-level graph classification problem and proposing HERMES, a heterogeneous edge-relational graph neural network with multi-head SSM-embedded attention. Vehicles and pedestrians are represented as heterogeneous nodes, while vehicle-vehicle, vehicle-pedestrian, and pedestrian-pedestrian interactions are modeled as relation-specific edges enriched with continuous, physics-informed descriptors. Relation-specific attention, dynamic node-edge updates, safety-aware graph pooling, and temporal sequence learning are jointly used to estimate scene-level conflict probability. HERMES was evaluated using 109,028 trajectory-derived sequences from a signalized urban intersection. Enhanced HERMES achieved an AUC-ROC of $0.9898 \pm 0.0013$, an AUC-PR of $0.9412 \pm 0.0067$, and an F1 score of $0.8449 \pm 0.0103$. At a 5% false-alarm rate, it detected 95.7% of conflict sequences, compared with 81.0% for the strongest Transformer baseline and 61.5% for XGBoost. External evaluation at an independently collected, broadly comparable intersection yielded a zero-shot AUC-ROC of 0.9752, an AUC-PR of 0.7829, and a sensitivity of 91.45% at a 5% false-alarm rate without target-site training. Joint source-target training using all source-site data and selected target-site segments further improved performance. Notably, inclusion of only 20% of the target-site training segments produced the highest low-false-alarm sensitivity, reaching $0.9869 \pm 0.0057$ at a 5% false-alarm rate, while the 60% setting achieved the highest AUC-ROC of $0.9920 \pm 0.0022$. These findings demonstrate that preserving heterogeneous interaction structure and short-term temporal evolution supports effective transfer across similar signalized-intersection environments and that substantial operational gains can be achieved with limited target-site data. HERMES therefore provides a promising foundation for low-false-alarm roadside conflict monitoring and future proactive traffic safety applications.

## 1 INTRODUCTION

Traffic safety at urban intersections remains a persistent public-health and transportation-engineering challenge. Road traffic crashes cause approximately 1.19 million deaths annually worldwide, with pedestrians and cyclists accounting for a disproportionate share of fatalities (World Health Organization, 2023). In the United States, 39,254 fatalities, approximately 2.42 million injuries, and 6.18 million police-reported crashes were recorded in 2024 (National Center for Statistics and Analysis, 2026). Pedestrian risk is particularly concentrated in urban environments: in 2023, pedestrians accounted for 7,314 fatalities and an estimated 68,244 injuries, representing 18% of all traffic fatalities; 84% of pedestrian fatalities occurred in urban areas and 17% at intersections (National Center for Statistics and Analysis, 2025). Signalized intersections remain critical safety locations because crossing, turning, yielding, queuing, and lane-changing maneuvers bring heterogeneous road users into close and rapidly changing interaction patterns.

Conventional safety analysis relies primarily on police-reported crashes. Although essential for long-term diagnosis, crash records are reactive, statistically sparse at individual intersections, and often lack the detailed behavioral and kinematic information needed to characterize how unsafe interactions develop immediately before a collision (Das, 2022; Hou et al., 2025; Hu et al., 2022). These limitations have motivated proactive approaches based on traffic conflicts and surrogate safety measures (SSMs). Measures such as time-to-collision (TTC), post-encroachment time (PET), deceleration rate to avoid a crash (DRAC), spacing, relative speed, and acceleration disparity quantify temporal and spatial proximity and occur sufficiently often to support more responsive safety assessment (Cooper, 1984; Formosa et al., 2020; Hayward, 1972). Advances in roadside video analytics, object detection, tracking, and trajectory extraction have further enabled continuous estimation of conflict-related states from observed traffic scenes (Formosa et al., 2020; Liu et al., 2026).

Machine-learning (ML) and deep-learning (DL) models have increasingly been used to classify or predict conflicts from traffic-flow variables, signal-cycle information, trajectories, and surrogate indicators (Hou et al., 2025; Zhang et al., 2024; Zhang and Abdel-Aty, 2022). However, signalized-intersection conflicts rarely arise from a single road-user pair or one critical TTC value. They often emerge from coupled interactions among through and turning vehicles, following traffic, pedestrians, and other nearby agents. Reliable scene-level assessment therefore requires a representation that preserves both interaction structure and its temporal evolution.

Existing methods remain limited in four respects. First, threshold-based SSM approaches classify a condition as hazardous when TTC, PET, or a related indicator crosses a predefined value. Their simplicity is attractive, but their rigid assumptions-often constant velocity or acceleration-do not adequately represent evasive behavior, surrounding traffic pressure, or contextual complexity(Chen et al., 2024; Lin et al., 2025). TTC-based alarms can also become unstable as the forecast horizon increases, particularly in heterogeneous pedestrian-rich environments with abrupt motion changes (Abdelraouf et al., 2022; De Ceunynck, 2017; Lin et al., 2025).

Second, supervised conflict models, including logistic regression (LR), support vector machines (SVM), random forests (RF), gradient boosting, recurrent networks, CNN-LSTM models, and Transformers typically encode each observation as a flat feature vector (Formosa et al., 2023; Hou et al., 2025; Zhang and Abdel-Aty, 2022). Although these models capture nonlinear relationships more effectively than fixed thresholds, flattening suppresses agent identity, geometric topology, and the dependency structure among simultaneous interactions. Consequently, scenes with different local conflict configurations may be represented by similar aggregated descriptors.

Third, many studies evaluate vehicle-vehicle (VV) or vehicle-pedestrian (VP) interactions independently (Arun et al., 2021; Chen et al., 2017; Lu et al., 2022; Muduli et al., 2026). This dyadic formulation cannot fully represent relational cascades in which one yielding, braking, or gap-acceptance decision alters the behavior and risk exposure of several surrounding agents. Pairwise sequence models are therefore structurally constrained when indirect multi-agent effects govern scene-level risk (Golchoubian et al., 2023; Muduli et al., 2026). Moreover, localization jitters, identity switches, missed detections, and occlusion can distort trajectory-derived TTC and PET estimates, making isolated scalar indicators particularly sensitive to perception error (Liu et al., 2026).

Fourth, graph neural networks (GNNs) can preserve relational structure, but existing graph-based safety models commonly use homogeneous agent assumptions or static adjacency structures that do not distinguish the semantics of VV, VP, and pedestrian-pedestrian (PP) interactions (Lyu et al., 2025; Sonth et al., 2025). SSMs are also generally appended as external predictors rather than embedded within the message-passing mechanism that determines how interaction information is propagated.

These limitations indicate that existing graph-based safety studies remain limited in jointly combining heterogeneous road-user relations, continuous surrogate-safety information, dynamic node-edge interaction modeling, short-term temporal encoding, and cross-site evaluation within a unified framework. Building on recent scene-graph and multi-relational traffic-safety models (Malazizi et al., 2026; Muduli et al., 2026; Sonth et al., 2025), this study formulates traffic conflict assessment as temporal scene-level heterogeneous graph classification and proposes HERMES, a Heterogeneous Edge-Relational Multi-Head Embedded SSM Attention framework. Vehicles and pedestrians are represented as heterogeneous nodes, while VV, VP, and PP interactions are modeled through relation-specific edges containing continuous kinematic and surrogate-safety descriptors. HERMES uses SSM-informed multi-head relational attention to modulate interaction importance, dynamic node-edge refinement to update relational representations during message passing, and temporal sequence encoding to characterize the short-term evolution of scene-level conflict states.

The study makes four principal contributions:

- It introduces a dynamic heterogeneous scene-graph framework that jointly models mixed vehicle-pedestrian interactions and updates node and edge representations as relational states evolve.
- It incorporates continuous safety-relevant descriptors directly into relation-specific multi-head attention, allowing kinematic and surrogate-safety information to influence the propagation of interaction messages.
- It combines heterogeneous graph learning with temporal sequence modeling to represent both within-frame multi-agent interaction structure and short-term changes across consecutive traffic scenes.
- It evaluates the framework through controlled comparisons with machine-learning, temporal deep-learning, and graph-based models, together with ablation, multi-seed robustness, low-false-alarm, and independent-site transfer analyses.

By integrating dynamic node-edge representation learning, SSM-informed relational attention, temporal encoding, and cross-site evaluation, HERMES provides a structured framework for scene-level conflict classification in mixed road-user environments at signalized intersections.

## 2 LITERATURE REVIEW

### 2.1 Surrogate Safety Measures for Conflict Identification

Traffic safety analysis has traditionally relied on police-reported crashes to identify hazardous locations and prioritize countermeasures. However, crash-based assessment is inherently reactive and statistically inefficient at the facility level because crashes are rare, random, and severe events that often require long observation periods to produce stable estimates (Hydén, 1987; Tarko and Lizarazo, 2021). This delays intervention until after serious harm has occurred, contrary to proactive safety principles such as Vision Zero (Kim et al., 2017). Surrogate safety measures (SSMs) support proactive assessment by quantifying near-misses, evasive maneuvers, and unsafe interactions from the temporal, spatial, and kinematic proximity of road users. The underlying premise is that conflicts and crashes share related failure mechanisms, while conflicts occur more frequently and therefore provide a richer basis for continuous risk monitoring (Arun et al., 2021; Tarko and Lizarazo, 2021). With trajectories increasingly available from video, drones, LiDAR, and connected-vehicle systems, SSMs can support infrastructure-based monitoring and intelligent intersection management.

Time-to-collision (TTC) is among the most widely used temporal indicators and estimates the remaining time to collision if current paths and speeds are maintained (Hayward, 1972). Its interpretability has encouraged widespread use, but its assumptions of constant speed, direction, and collision geometry are often violated at urban intersections, where turning, yielding, braking, and pedestrian path changes are common. TTC may therefore generate false alarms when evasive action is already underway or miss emerging hazards that are not reflected in the current trajectory state (Chen et al., 2024; Lin et al., 2025).

Post-encroachment time (PET) measures the interval between one road user leaving a conflict point and another arriving at the same location (Cooper, 1984). PET is useful for retrospective diagnosis of spatially overlapping trajectories, but because it is calculated after one user has cleared the conflict area, it is less suitable for prospective warning. Deceleration rate to avoid a crash (DRAC) instead estimates the minimum deceleration required to avert a collision and captures evasive demand more directly (Alhajyaseen, 2015). Additional measures, including proportion of stopping distance, gap time, relative speed, spacing, and acceleration disparity, provide complementary information on maneuvering space and response intensity (Kathuria and Vedagiri, 2020; Lyu et al., 2025). More recent formulations have relaxed conventional conflict-point assumptions; for example, Lin et al. (2025) proposed predicted lane-based TTC using a lane-level conflict zone and deep-learning-based arrival-time estimation.

Evidence increasingly favors combining multiple SSMs rather than relying on a single threshold. Kathuria and Vedagiri (2020) combined speed, TTC, gap time, and PET to characterize pedestrian-vehicle interaction severity. Amini et al. (2022) integrated time-, distance-, and speed-based indicators to better represent evasive behavior, while Chen et al. (2024) showed that TTC performance deteriorated when the prediction horizon exceeded approximately two seconds. These findings indicate that SSMs are informative but sensitive to context, behavioral adaptation, and forecast horizons. Their reliability also depends on trajectory quality. Detection errors, occlusion, localization jitter, and identity switches can propagate directly into SSM estimates. Liu et al. (2026) found that automatically derived PET was generally more robust than TTC, but both remained less reliable than measures computed from ground-truth trajectories. Because small positional perturbations can substantially alter TTC, PET, and DRAC, isolated scalar thresholds or flat SSM inputs may not adequately represent complex intersection risk.

### 2.2 Machine Learning Approaches for Conflict Prediction

ML methods have been widely adopted to overcome the rigidity of fixed SSM thresholds. Rather than identifying conflicts only when TTC, PET, or DRAC crosses a predefined value, ML models learn nonlinear associations between conflict occurrence and variables such as traffic volume, speed, acceleration, queue length, signal phase, pedestrian exposure, spacing, relative speed, and surrogate indicators. Statistical models, including logistic regression (LR), generalized linear models, safety performance functions, and extreme-value approaches, remain valuable for interpretation but often require assumptions concerning functional form, distribution, and independence (Essa and Sayed, 2018; Hou et al., 2025; Hu et al., 2022; Mukherjee and Mitra, 2019; Zheng and Sayed, 2020). By contrast, SVM, decision trees (DT), RF, gradient boosting, and XGBoost can accommodate nonlinear and correlated predictors with fewer parametric assumptions (Hou et al., 2025; Hu et al., 2022).

Empirical studies demonstrate the value of these methods for proactive safety assessment. Hu et al. (2022) compared LR with several ML classifiers using high-resolution trajectory data and reported an accuracy of 0.85 for RF. Zhang and Abdel-Aty (2022) combined CCTV-derived TTC and PET, signal-performance measures, and pedestrian-exposure variables for signal-cycle-level pedestrian conflict prediction; their XGBoost model achieved an Area Under the Curve (AUC) of 0.841 and recall of 0.739. ML has also supported related behavioral tasks, such as predicting vehicle-yielding intentions from speed, crosswalk distance, pedestrian presence, preceding-vehicle behavior, and surrounding traffic context (Muduli and Ghosh, 2024). Collectively, these studies show that conflict risk depends on both kinematic conditions and behavioral context.

Three limitations remain. First, conflict datasets are strongly imbalanced because safe interactions outnumber hazardous events. Consequently, high accuracy may conceal poor minority-class detection. Formosa et al. (2023) emphasized the use of sensitivity, false-alarm rate, AUC, and Brier score, together with cost-sensitive learning and threshold adjustment, for more reliable evaluation. Second, most ML models operate on flat feature vectors. Although such vectors may contain TTC, distance, relative speed, and DRAC, they do not preserve which agents generated these values or how those agents interact with the surrounding scene. This loss of relational identity limits the representation of multi-agent dependencies and risk propagation. Third, transferability remains problematic because geometry, traffic composition, signal timing, camera perspective, and road-user behavior vary across locations. Hou et al. (2025) noted that models trained at one site may experience substantial degradation when applied elsewhere. Thus, although ML improves upon fixed-threshold approaches, its effectiveness remains constrained by class imbalance, flat scene representation, and limited cross-site generalizability.

### 2.3 Deep Learning Approaches for Conflict Prediction

Deep-learning (DL) methods have been used to capture nonlinear feature interactions and temporal dependencies in conflict prediction. Unlike conventional ML models, DL architecture can process disaggregated trajectories, image-derived features, sensor streams, and time-series traffic variables, making them suitable for short-horizon safety assessment (Formosa et al., 2020; Hou et al., 2025; Zhang et al., 2024).

Formosa et al. (2020) developed an real-time conflict-prediction framework that integrated in-vehicle sensing, traffic conditions, weather, speed, density, and multiple SSMs within a deep neural network. The study demonstrated that conflict risk cannot be characterized adequately through TTC thresholds alone and that heterogeneous traffic and environmental variables can improve proactive safety assessment. Subsequent work compared DNN, LSTM, and CNN-LSTM architectures with conventional classifiers under class imbalance, showing that temporal DL models can improve minority conflict detection when imbalance is addressed appropriately (Formosa et al., 2023).

Recurrent models are particularly relevant because conflict risk evolves through changes in speed, acceleration, spacing, and surrogate-safety states. More recent hybrid and attention-based models have extended this capability. Hou et al. (2025) proposed a Gated-Transformer for merging-area conflict prediction and reported an F1 score of 0.864 and an AUC of 0.980, while also demonstrating the value of transfer learning for cross-site applications. Zhang et al. (2025) combined video-derived trajectories, lane-level traffic variables, a Deep and Cross Network, and SHAP analysis to capture interactions among traffic flow, speed, and platoon characteristics. Transformer-based models have also been applied to yielding-intention prediction, identifying pedestrian visibility, time to the crosswalk, and preceding-vehicle behavior as influential factors (Muduli and Ghosh, 2024).

Despite these advances, most DL models continue to represent traffic scenes as flat temporal matrices. An LSTM may learn the evolution of TTC, and a Transformer may attend across input tokens, but neither inherently preserves which agents generated a given interaction or how that interaction relates to the surrounding scene. Consequently, temporal dependence can be learned without explicitly representing the physical structure of multi-agent influence.

Trajectory-forecasting models introduce an additional limitation because small position errors can propagate across the prediction horizon and distort subsequent conflict estimates, particularly in pedestrian-rich environments with abrupt behavioral changes (Golchoubian et al., 2023; Lin et al., 2025). DL models are also sensitive to detection, tracking, and localization errors in automated video pipelines, which can degrade trajectory-derived safety indicators (Liu et al., 2026). These limitations motivate models that combine temporal learning with explicit relational structure and physics-informed edge representations rather than treating pairwise SSMs as independent scalar inputs.

### 2.4 Graph Neural Network Approaches for Traffic Safety

Graph neural networks (GNNs) provide a relational alternative to flat ML and sequence-based DL models by representing traffic participants, infrastructure, and roadway elements as nodes connected through interaction, proximity, right-of-way, or conflict edges. This formulation is well suited to traffic safety because risk emerges from direct and indirect dependencies among multiple agents rather than from isolated road-user states. Through message passing, GNNs can preserve these dependencies and learn scene-level representations of evolving traffic interactions.

Sonth et al. (2025) developed a semantic scene graph framework for intersection safety using real-world video data and TransformerConv and GINEConv architectures. Their model supported scenario-level risk classification, node-level threat assessment, and edge-level interaction analysis, achieving 79.8% accuracy. Muduli et al. (2026) proposed a multi-relational spatiotemporal GNN for pedestrian-vehicle safety at unsignalized crosswalks. By explicitly distinguishing PP, VV, and VP relations, their model avoided preselecting individual pairs and captured broader scene interactions. It achieved 90.6% accuracy, an F1 score of 0.906, and an AUC of 0.950, outperforming GRU, LSTM, and Transformer-based baselines.

Graph-based safety modeling has also been extended to other traffic environments. Lyu et al. (2025) combined graph attention with a causally structured covariate representation for conflict-risk prediction at expressway diversion zones, linking predictive modeling with interpretable risk relationships. In traffic decision-making, Wang et al. (2025) integrated a relational graph attention network with deep reinforcement learning for freeway weaving, while Tong et al. (2025) incorporated roadway topology into graph-based multi-agent intersection control. Zheng et al. (2025) similarly used graph attention and deep reinforcement learning to model explicit and implicit interaction dependencies at unsignalized intersections. Although these studies primarily address control and decision-making, they demonstrate the value of graph reasoning for multi-agent traffic systems.

Despite this progress, several limitations remain. Many existing models use homogeneous graphs, static adjacency structures, or undifferentiated interaction semantics. Others employ SSMs only as labels, targets, or external predictors rather than embedding them within graph edges and message passing. As a result, the graph may preserve topology without explicitly prioritizing safety-critical interactions. This gap motivates HERMES, which combines heterogeneous relation types with physics-informed edge descriptors so that relational attention is guided by both interaction structure and surrogate-safety information.

**2.5 Research Gaps and Positioning of HERMES**

The reviewed literature reveals five unresolved methodological gaps. First, most SSM-based approaches remain pairwise, evaluating VV or VP interactions independently. Although useful for localized risk screening, this formulation does not represent the coupled multi-agent dynamics through which conflicts emerge at signalized intersections.

Second, most ML and DL models flatten SSMs into feature vectors. Once TTC, distance, DRAC, relative speed, and related descriptors are aggregated, the model no longer preserves which agents generated them, what interaction type they represent, or how the pair relates to surrounding road users. This weakens the physical meaning of the safety indicators and suppresses scene-level dependency structure.

Third, temporal DL models such as LSTM, GRU, CNN-LSTM, and Transformer architectures can learn sequential patterns but do not inherently encode explicit relational topology. Consequently, they may capture temporal variation without representing how direct and indirect influences propagate across multiple interacting agents.

Fourth, although graph-based safety models preserve scene structure, most do not embed continuous surrogate-safety descriptors within graph edges and attention mechanisms. Existing approaches often define connectivity using distance, roadway topology, or interaction type, but rarely allow DRAC, relative speed, heading difference, distance, and relative velocity to directly govern message importance. This leaves a gap between surrogate-safety theory and graph-based relational learning.

Fifth, the literature lacks controlled comparisons across ML, temporal DL, and GNN models using identical data, feature sets, labels, and evaluation protocols. Without such benchmarking, it is difficult to determine whether reported gains arise from model architecture or from differences in data construction and experimental design.

HERMES addresses these limitations by formulating conflict assessment as temporal scene-level heterogeneous graph classification. Vehicles and pedestrians are represented as nodes, while VV, VP, and PP interactions are modeled as typed edges. Continuous safety-relevant descriptors are embedded directly within the edges and used to guide multi-head attention, enabling the model to preserve relational identity and prioritize physically critical interactions.

The framework further incorporates dynamic node-edge refinement, allowing updated agent states to modify edge criticality and refined edge states to influence node representations. This reciprocal updating reflects the fact that conflict risk emerges jointly from road-user behavior and interaction dynamics. Finally, HERMES is evaluated through a controlled three-tier benchmark across ML, temporal DL, and graph-based models, providing a direct test of whether relational graph learning exploits SSM information more effectively than flat alternatives.

# 3 METHODOLOGY

The methodological framework transforms roadside video into structured spatiotemporal graphs for scene-level traffic conflict classification. As summarized in Figure 1, the workflow comprises five stages: data acquisition and trajectory extraction, trajectory preprocessing, kinematic and surrogate-safety feature construction, heterogeneous graph generation, and comparative model evaluation.

First, fixed-camera video data are processed using road-user detection and multi-object tracking to obtain vehicle and pedestrian trajectories. The trajectories are then refined through region-of-interest filtering, image-to-ground coordinate transformation, quality control, and smoothing to reduce localization noise and support reliable microscopic safety analysis.

Second, the processed trajectories are used to derive node-level motion variables and interaction-level safety descriptors. Speed, acceleration, and heading characterize individual road-user motion, while distance, relative speed, time-to-collision, post-encroachment time, deceleration rate to avoid a crash, and related measures describe interaction proximity and evasive demand. Pairwise conflict information is subsequently aggregated into frame- and sequence-level binary labels.

Third, each frame is represented as a heterogeneous scene graph. Vehicles and pedestrians form distinct node types, while VV, VP, and PP interactions are encoded as typed edges. Safety-relevant interaction descriptors are embedded as edge attributes, preserving both relational identity and traffic-scene topology.

Finally, the resulting graph sequences are processed by HERMES, which combines multi-head SSM-informed attention, dynamic node-edge refinement, safety-aware graph pooling, temporal sequence encoding, and binary classification. Performance is compared with conventional ML, temporal DL, and graph-based baselines under consistent data splits, input construction, and evaluation protocols. The assessment includes global discrimination, low-false-alarm performance, confusion-matrix analysis, multi-seed robustness, ablation testing, and cross-site transfer evaluation.

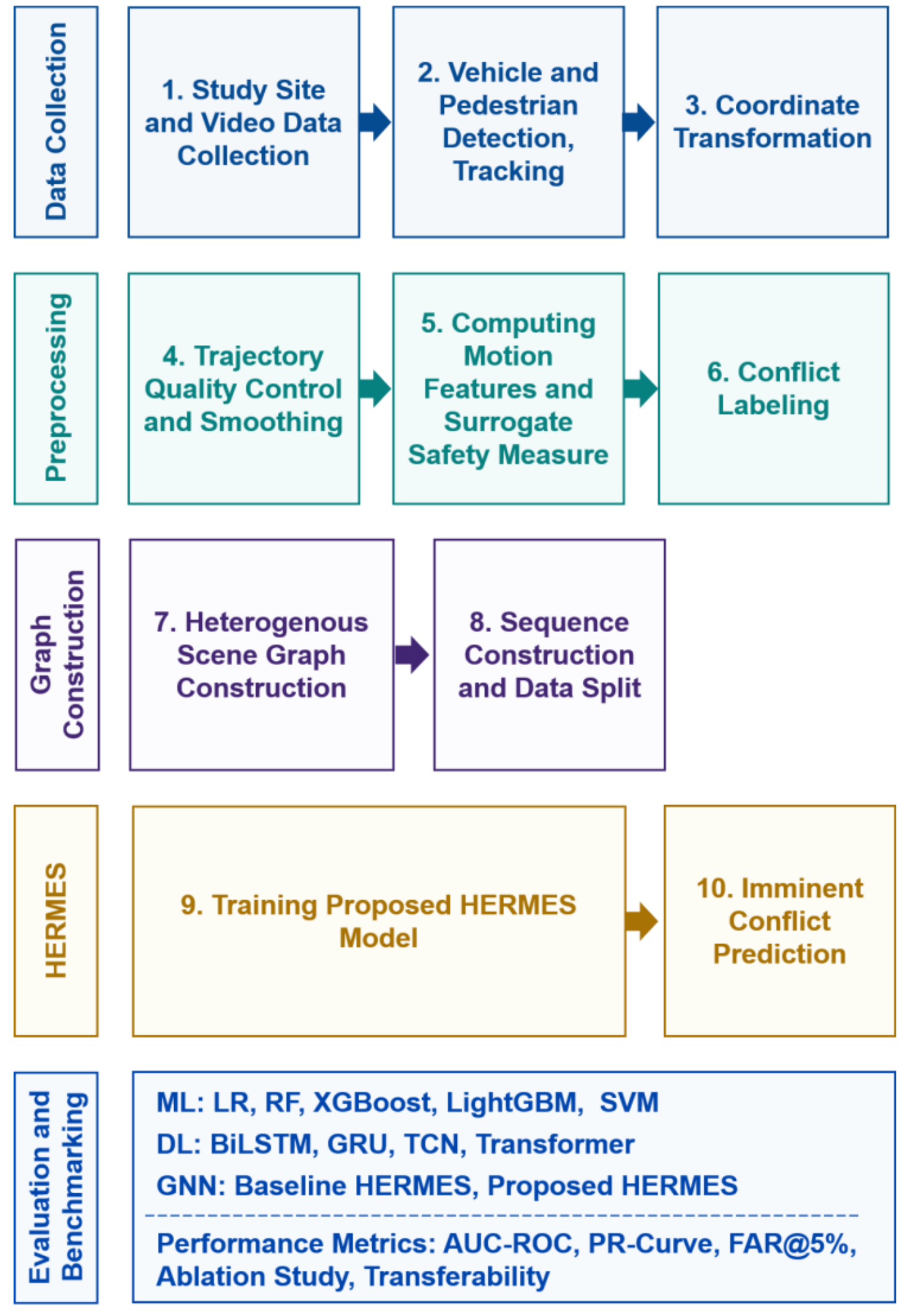

Figure 1: Overall methodological workflow of the proposed HERMES framework.

### 3.1 HERMES Model Architecture

The proposed HERMES framework maps a temporally ordered sequence of heterogeneous traffic-scene graphs to a scene-level conflict prediction. Each frame is represented as a unified relational graph containing all observed vehicles and pedestrians, allowing the model to capture both within-frame interaction topology and short-term risk evolution across consecutive frames.

For each frame, two heterogeneous graph propagation layers apply relation-specific multi-head SSM-embedded attention, relation fusion, residual node updating, and dynamic edge refinement. The resulting node representations are aggregated through DRAC-guided graph pooling. The sequence of frame-level embeddings is then processed by a two-layer LSTM, and the final temporal representation is mapped to a scene-level conflict logit. Figure **2** illustrates the architecture, while Algorithm 1 summarizes the complete forward-propagation procedure. The Multi-Head SSM Attention and Edge Update operators are detailed in Algorithms 2 and 3, respectively.

---

**Algorithm 1:** Forward Propagation of the HERMES Framework

---

**Input:** Temporal heterogeneous graph sequence

$S = \{\mathbf{X}_t, \mathbf{A}_t^{VV}, \mathbf{A}_t^{VP}, \mathbf{A}_t^{PP}, \mathbf{E}_t\}_{t=1}^{T}$

**Output:** Scene-level conflict logit $\hat{\mathbf{y}}$

1 $Initialize frame - embedding\ list\ \mathcal{F} \leftarrow [\,]$

2 **for** $t = 1, \dots, T$ **do**

3 $\quad H_t^{(0)} \leftarrow \mathrm{ReLU}\big(\mathrm{InputProj}(X_t)\big)$

4 $\quad E_t^{(0)} \leftarrow E_t$

5 $\quad A_t^{*} \leftarrow \mathrm{clamp}(A_t^{VV} + A_t^{VP} + A_t^{PP}, 0,1)$

6 $\quad$ **for** $l = 1$ to $2$ **do**

7 $\quad\quad M_{t,VV}^{(l)} \leftarrow \mathrm{MHSSMAttn}_{VV}^{(l)}\left(H_t^{(l-1)}, A_t^{VV}, E_t^{(l-1)}\right)$

8 $\quad\quad M_{t,VP}^{(l)} \leftarrow \mathrm{MHSSMAttn}_{VP}^{(l)}\left(H_t^{(l-1)}, A_t^{VP}, E_t^{(l-1)}\right)$

9 $\quad\quad M_{t,PP}^{(l)} \leftarrow \mathrm{MHSSMAttn}_{PP}^{(l)}\left(H_t^{(l-1)}, A_t^{PP}, E_t^{(l-1)}\right)$

10 $\quad\quad C_t^{(l)} \leftarrow H_t^{(l-1)} \parallel M_{t,VV}^{(l)} \parallel M_{t,VP}^{(l)} \parallel M_{t,PP}^{(l)}$

11 $\quad\quad \widetilde{H}_t^{(l)} \leftarrow \mathrm{ReLU}\left(\mathrm{Fusion}^{(l)}\left(C_t^{(l)}\right)\right)$

12 $\quad\quad H_t^{(l)} \leftarrow H_t^{(l-1)} + \mathrm{LayerNorm}_h^{(l)}\left(\widetilde{H}_t^{(l)}\right)$

13 $\quad\quad E_t^{(l)} \leftarrow \mathrm{EdgeUpdate}^{(l)}\left(H_t^{(l)}, E_t^{(l-1)}, A_t^{*}\right)$

14 $\quad$ **end for**

15 $\quad \overline{H}_t \leftarrow \mathrm{Dropout}\left(H_t^{(2)}\right)$

16 $\quad$ **for** $i = 1$ to $N$ **do**

17 $\quad\quad s_{t,i} \leftarrow \max_{j=1,\dots,N} E_{t,ij,k_{\mathrm{DRAC}}}^{(2)}$

18 $\quad$ **end for**

19 $\mathbf{w}_t \leftarrow \text{Softmax}(\mathbf{s}_t)$

20 $g_t \leftarrow \sum_{i=1}^{N} w_{t,i} \overline{H}_{t,i}$

21 $\text{Append}(\mathcal{F}, g_t)$

22 **end for**

23 $H_{\text{seq}} \leftarrow \text{Stack}(g_1, g_2, \dots, g_T)$

24 $Z_{:,-} \leftarrow \text{LSTM}(H_{\text{seq}})$

25 $z_T \leftarrow \text{Dropout}(Z[:, T, :])$

26 $\hat{y} \leftarrow \text{Squeeze}(\text{FC}(z_T))$

27 **return** $\hat{\boldsymbol{y}}$

Heterogeneous Scene-Graph Representation

For observation frame *t*, the heterogeneous traffic-scene graph is represented as:

$$G_t = \left(X_t, A_t^{vv}, A_t^{vp}, A_t^{pp}, E_t\right) \tag{1}$$

where $X_t \in \mathbb{R}^{N \times F_n}$ is the node-feature matrix, $A_t^r \in \{0,1\}^{N \times N}$ is the adjacency matrix associated with relation $r \in \{\text{VV}, \text{VP}, \text{PP}\}$, and $E_t \in \mathbb{R}^{N \times N \times F_e}$ is the edge-feature tensor. The VV, VP, and PP adjacency matrices respectively represent vehicle-vehicle, vehicle-pedestrian, and pedestrian-pedestrian relations. Here, $N$ denotes the fixed maximum padded number of agents represented within a frame.

The raw node features are projected into a latent hidden space:

$$H_t^{(0)} = \text{ReLU}\ (X_t W_{\text{in}} + b_{\text{in}}) \tag{2}$$

where $W_{\text{in}}$ is a learnable input projection matrix and $b_{\text{in}}$ is a bias parameter. Unlike a homogeneous graph encoder that treats all interactions uniformly, HERMES performs message aggregation separately over the VV, VP, and PP channels. This preserves the distinct semantics associated with different road-user interaction types.

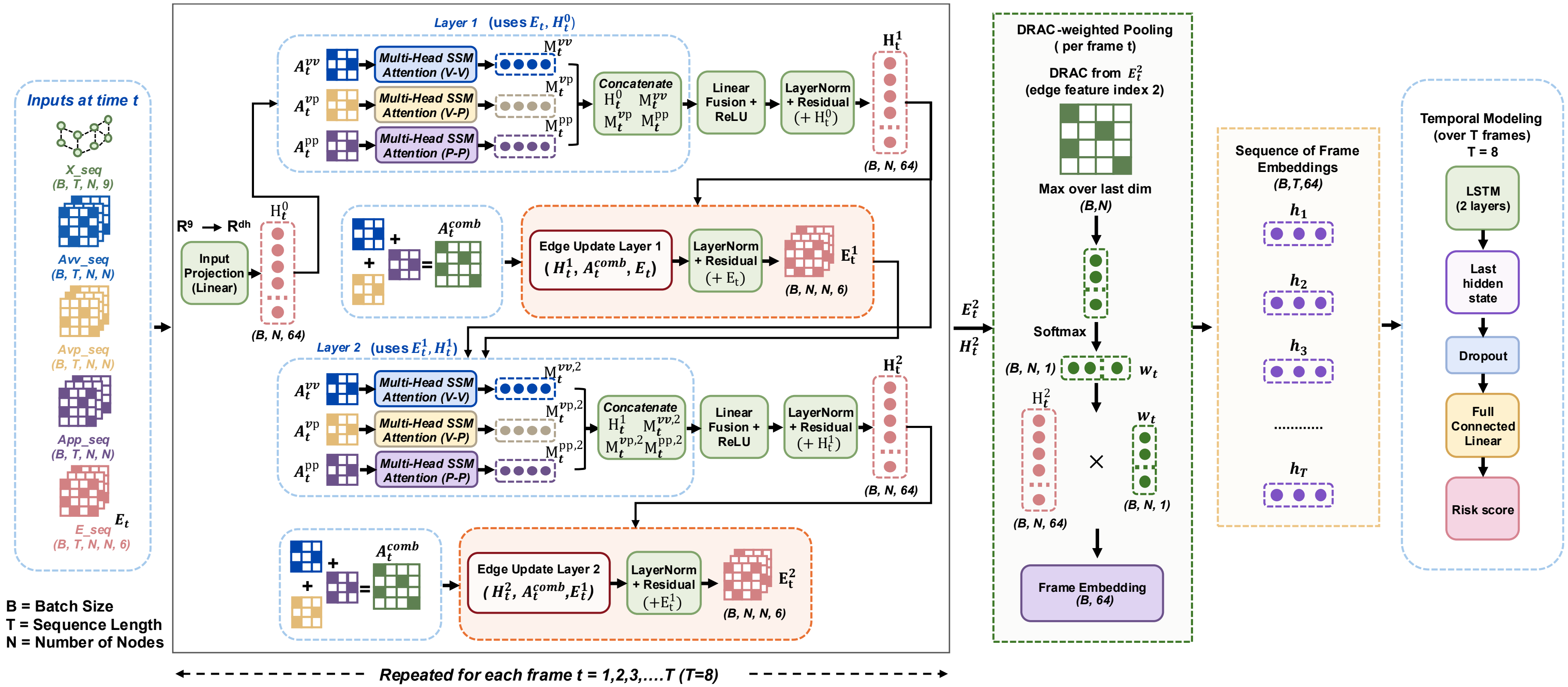

Figure 2: Architecture of the proposed HERMES framework for conflict risk prediction.

Relation-Specific Multi-Head SSM-Embedded Attention

At graph propagation layer $\ell$, a separate multi-head attention module is applied to each relation type $r \in \{\mathrm{VV}, \mathrm{VP}, \mathrm{PP}\}$. The query, key, and value projections are computed as:

$$Q_t^{r,\ell} = H_t^{(\ell-1)} W_q^{r,\ell} + b_q^{r,\ell} \tag{3}$$

$$K_t^{r,\ell} = H_t^{(\ell-1)} W_k^{r,\ell} + b_k^{r,\ell} \tag{4}$$

$$V_t^{r,\ell} = H_t^{(\ell-1)} W_v^{r,\ell} + b_v^{r,\ell} \tag{5}$$

The projected representations are partitioned into $K_h$ attention heads, where the dimension of each head is $d_h = d/K_h$. The edge representation associated with node pair $(i, j)$ is transformed into one learned attention bias for each head:

$$B_{t,h,ij}^{r,\ell} = \left[\mathrm{EdgeEnc}_{r,\ell}\left(E_{t,ij}^{(\ell-1)}\right)\right]_h \tag{6}$$

The unnormalized attention score for head $h$ is derived by incorporating the learned edge bias:

$$S_{t,h,ij}^{r,\ell} = \frac{\left(q_{t,h,i}^{r,\ell}\right)^{\top} k_{t,h,j}^{r,\ell}}{\sqrt{d_h}} + B_{t,h,ij}^{r,\ell} \tag{7}$$

The relation-specific adjacency matrix is used to mask invalid connections before normalization:

$$\tilde{S}_{t,h,ij}^{r,\ell} = \begin{cases} S_{t,h,ij}^{r,\ell}, & A_{t,ij}^{r} = 1, \\ -\infty, & A_{t,ij}^{r} = 0. \end{cases} \tag{8}$$

In the computational implementation, the masked entries are assigned a large negative constant. The attention coefficients are normalized over the destination-node dimension:

$$\alpha_{t,h,ij}^{r,\ell} = \mathrm{softmax}_j\left(\tilde{S}_{t,h,ij}^{r,\ell}\right) \tag{9}$$

The head-specific value representations are aggregated as:

$$O_{t,h,i}^{r,\ell} = \sum_j \alpha_{t,h,ij}^{r,\ell} \, V_{t,h,j}^{r,\ell} \tag{10}$$

The outputs of all attention heads are concatenated and linearly projected:

$$M_t^{r,\ell} = \mathrm{OutProj}_{r,\ell}\left[\mathrm{Concat}_{h=1}^{K_h}\left(O_{t,h,i}^{r,\ell}\right)\right] \tag{11}$$

Because the attention score explicitly includes a learned function of the edge-level surrogate safety descriptors, the importance of each interaction is determined jointly by its latent node compatibility and its physical safety characteristics. Multiple attention heads allow the network to learn complementary relational patterns associated with relative motion, directional convergence, spacing, and braking demand. Algorithm 2 summarizes the relation-specific attention operation.

**Algorithm 2:** Relation-Specific Multi-Head SSM-Embedded Attention

| | |
|---|---|
| **Input:** | Node representations $H$, relation-specific adjacency matrix $A^r$, and edge representations $E$ |
| **Output:** | Relation-specific node messages $M^r$ |
| 1 | Set the number of attention heads to $K_h$ and $d_h \leftarrow d/K_h$ |
| 2 | $Q \leftarrow HW_q^r + b_q^r$ |
| 3 | $K_r \leftarrow HW_k^r + b_k^r$ |
| 4 | $V \leftarrow HW_v^r + b_v^r$ |
| 5 | Partition and rearrange $Q$, $K_r$, and $V$ into $K_h$ attention heads |
| 6 | $S \leftarrow QK_r^\top/\sqrt{d_h}$ |
| 7 | $B_E \leftarrow W_e^r E + b_e^r$ |
| 8 | Rearrange $B_E$ to align the edge biases with the attention heads |
| 9 | $S \leftarrow S + B_E$ |
| 10 | Replicate $A^r$ across the attention heads |
| 11 | Set $S_{hij} \leftarrow -10^9$ wherever $A_{ij}^r = 0$ |
| 12 | $\alpha \leftarrow \mathrm{Softmax}_j(S)$ |
| 13 | $O \leftarrow \alpha V$ |
| 14 | Rearrange and concatenate the outputs of all attention heads |
| 15 | $M^r \leftarrow W_o^r O + b_o^r$ |
| 16 | **return $\boldsymbol{M^r}$** |

Relation Fusion and Residual Node Update

The outputs of the three relation-specific attention modules are concatenated with the node representation from the preceding layer:

$$C_t^{(\ell)} = H_t^{(\ell-1)} \| M_t^{\mathrm{VV},\ell} \| M_t^{\mathrm{VP},\ell} \| M_t^{\mathrm{PP},\ell} \tag{12}$$

where $\|$ denotes feature-wise concatenation. The concatenated representation is transformed as:

$$\widetilde{H}_t^{(\ell)} = \mathrm{ReLU}\left(C_t^{(\ell)} W_f^{(\ell)} + b_f^{(\ell)}\right) \tag{13}$$

The node state is updated using layer normalization and a residual connection:

$$H_t^{(\ell)} = \mathrm{LayerNorm}_{h,\ell}\left(\widetilde{H}_t^{(\ell)}\right) + H_t^{(\ell-1)} \tag{14}$$

The residual connection preserves the preceding node state while allowing each graph propagation layer to introduce relation-specific interaction information.

Dynamic Edge Refinement

Static edge features cannot reflect changes in relational importance as the latent node states evolve. HERMES therefore updates the edge representations after each node-update stage. A combined adjacency mask is first constructed as:

$$A_t^* = \text{clamp}\left(A_t^{\text{VV}} + A_t^{\text{VP}} + A_t^{\text{PP}}, 0{,}1\right) \tag{15}$$

For each ordered node pair $(i, j)$, a candidate edge representation is computed by concatenating the updated source-node state, the updated destination-node state, and the preceding edge representation:

$$\tilde{E}_{t,ij}^{(\ell)} = \text{ReLU}\left[W_e^{(\ell)}\left(H_{t,i}^{(\ell)} \| H_{t,j}^{(\ell)} \| E_{t,ij}^{(\ell-1)}\right) + b_e^{(\ell)}\right] \tag{16}$$

The edge representation is then updated with a layer-normalized residual step according to:

$$E_{t,ij}^{(\ell)} = \text{LayerNorm}_{e,\ell}\left[E_{t,ij}^{(\ell-1)} + A_{t,ij}^*\left(\tilde{E}_{t,ij}^{(\ell)} - E_{t,ij}^{(\ell-1)}\right)\right] \tag{17}$$

When $A_{t,ij}^* = 1$, the candidate edge representation replaces the preceding representation before layer normalization. When $A_{t,ij}^* = 0$, the preceding representation is retained. The refined edge tensor is then passed to the next graph-propagation layer, allowing updated node states to influence subsequent relation-specific attention. Algorithm 3 summarizes this procedure.

| **Algorithm 3:** Dynamic Edge Update | |
|---|---|
| **Input:** | Node representations $H$, preceding edge representations $E$, and combined adjacency matrix $A^*$ |
| **Output:** | Updated edge representations $E'$ |
| 1 | Expand $H$ along the destination-node dimension to obtain $H_i$ |
| 2 | Expand $H$ along the source-node dimension to obtain $H_j$ |
| 3 | $C_{ij} \leftarrow H_i \parallel H_j \parallel E_{ij}$ |
| 4 | $\tilde{E}_{ij} \leftarrow \text{ReLU}\left(W_{\text{edge}} C_{ij} + b_{\text{edge}}\right)$ |
| 5 | Expand $A^*$ along the edge-feature dimension to obtain $\mathcal{M}$ |
| 6 | $E_{ij}^{\text{masked}} \leftarrow E_{ij} + \mathcal{M}_{ij}\left(\tilde{E}_{ij} - E_{ij}\right)$ |
| 7 | $E'_{ij} \leftarrow \text{LayerNorm}\left(E_{ij}^{\text{masked}}\right)$ |
| 8 | **return** $\boldsymbol{E'}$ |

DRAC-Guided Graph Pooling
After the second heterogeneous graph propagation layer, dropout is applied to the final node representations:

$$\overline{H}_t^{(2)} = \text{Dropout}\left(H_t^{(2)}\right) \quad (18)$$

The DRAC channel of the updated edge tensor is used to determine the relative contribution of each node to the frame-level representation. For node $i$, the local risk score is defined as the maximum DRAC value among its outgoing edges.

$$s_{t,i} = \max_j E_{t,ij,k_{\text{DRAC}}}^{(2)} \quad (19)$$

where $k_{\text{DRAC}}$ identifies the target DRAC position channel inside the edge feature matrices. The risk scores are normalized across the nodes via a row-wise softmax layout:

$$w_{t,i} = \frac{\exp\left(s_{t,i}\right)}{\sum_{k=1}^{N} \exp\left(s_{t,k}\right)} \quad (20)$$

The frame-level graph representation is then computed as:

$$g_t = \sum_{i=1}^{N} w_{t,i} \overline{H}_{t,i}^{(2)} \quad (21)$$

This risk-aware pooling operation assigns greater influence to agents associated with high braking-demand interactions and reduces the contribution of less critical background objects to the scene-level representation.

Temporal Encoding and Scene-Level Prediction
The frame-level graph embeddings are chronologically stacked:

$$H_{\text{seq}} = \text{Stack}(g_1, g_2, \dots, g_T) \quad (22)$$

The resulting matrix sequence is processed by the LSTM temporal encoder:

$$Z, _ = \text{LSTM}\left(H_{\text{seq}}\right) \quad (23)$$

where $Z$ contains the LSTM output at every observation time step. The hidden state vector corresponding to the terminal frame is passed through dropout regularization and a fully connected classification layer:

$$\hat{y} = W_c \, \text{Dropout}(Z[:, T, :]) + b_c \quad (24)$$

where $\hat{y}$ denotes the scene-level conflict logit. The model outputs this raw logit during forward propagation, and the corresponding conflict probability is obtained during inference using the sigmoid function:

$$P(\text{conflict} \mid \mathcal{S}) = \sigma(\hat{y}) = \frac{1}{1 + \exp(-\hat{y})} \tag{25}$$

The model thus maps a temporally ordered heterogeneous graph sequence to a unified scene-level conflict probability.

End-to-End Optimization

All HERMES components, including relation-specific attention, relation fusion, dynamic edge updating, DRAC-guided pooling, the LSTM encoder, and the classification layer, are jointly optimized using scene-level logits and binary conflict labels. Gradients propagate through both intra-frame graph operations and inter-frame temporal encoding, enabling simultaneous learning of relational interaction structure and its temporal evolution. Residual connections, layer normalization, nonlinear activation, and dropout are used to stabilize training and reduce overfitting.

The implementation uses sequences of $T = 8$ frames, a maximum of $N = 9$ nodes per frame, nine node features, and six edge features. The graph encoder comprises two message-passing layers with four attention heads and a hidden dimension of 64. DRAC occupies edge-feature index 2 under zero-based tensor indexing, corresponding to the third channel in the mathematical notation.

### 3.2 Baseline and Comparison Models

HERMES was evaluated using a three-tier benchmark comprising conventional ML, temporal DL, and graph-based models. All models used the same sequence-level binary conflict labels, while the input representation was adapted to each model family. The ML models received fixed-dimensional summary vectors derived from the complete 8-frame sequence, the DL models received temporally ordered frame-level feature vectors, and the graph models processed the full heterogeneous scene-graph sequence without flattening. This design isolates the contribution of statistical learning, temporal modeling, relational graph reasoning, and the SSM-informed components introduced in HERMES.

Machine Learning Tier

The ML tier includes five tabular classifiers: LR, RF, XGBoost, LightGBM, and support vector machine. These models evaluate whether scene-level conflict patterns can be identified from statistical descriptors without explicitly preserving temporal order or relational topology.

Because conventional ML models require fixed-length inputs, each 8-frame sequence was summarized using node-level motion variables, edge-level surrogate-safety measures, graph-structural characteristics, and temporal trend indicators. For a variable set $q$, the statistical summary operator was defined as:

$$S(q) = [\mu(q), \min(q), \max(q), \sigma(q)] \tag{26}$$

where $\mu(q)$, $\min(q)$,$\max(q)$, and $\sigma(q)$ denote the mean, minimum, maximum, and standard deviation computed over the complete sequence. This operator was used to construct the eleven feature groups summarized in Table 1.

Table 1: Extracted graph-level feature groups.

| Feature level | Feature group | Description |
|---|---|---|
| Node | Speed | Distribution of road-user speeds |
| Node | Acceleration | Distribution of acceleration magnitudes |
| Edge | Distance | Distribution of pairwise interaction distances |
| Edge | Relative speed | Distribution of pairwise relative speeds |
| Edge | DRAC | Distribution of required collision-avoidance deceleration |
| Edge | Heading difference | Distribution of absolute pairwise heading differences |
| Edge | Relative $v_x$ | Distribution of relative velocity along the $x$-axis |
| Edge | Relative $v_y$ | Distribution of relative velocity along the $y$-axis |
| Frame | Node count | Number of road-user nodes |
| Frame | Edge count | Number of directed interaction edges |
| Frame | Vehicle fraction | Proportion of nodes classified as vehicles |

The pooled sequence-level descriptors were supplemented with three temporal trend variables:

$$\Delta speed = speed_{last} - speed_{first} \tag{27}$$

$$\Delta DRAC = DRAC_{last} - DRAC_{first} \tag{28}$$

$$\Delta distance = distance_{last} - distance_{first} \tag{29}$$

Thus, each temporal scene sequence was represented by a fixed ML input vector:

$$z^{ML} \in \mathbb{R}^{39} \tag{30}$$

as implemented in the sequence-level feature extraction pipeline.

*Logistic Regression (LR)*

LR provides the simplest linear probabilistic baseline and evaluates whether conflict scenes are linearly separable within the engineered feature space. The predicted conflict probability is computed as:

$$P(\text{conflict}) = \sigma(w^T z^{ML} + b) \tag{31}$$

where $w$ is the learned coefficient vector, $b$ is the bias term, and $\sigma(\cdot)$ denotes the sigmoid function. Because LR assumes linear additive influence of the descriptors, it serves as a low-capacity benchmark against which more flexible nonlinear classifiers can be compared.

*Random Forest (RF)*

RF is a bagging-based tree ensemble that combines the predictions of multiple independently trained decision trees introduced by Breiman (2001). The final classification is obtained through majority voting:

$$\hat{y} = \text{mode}T_1(z^{ML}), T_2(z^{ML}), \ldots, T_K(z^{ML}) \tag{32}$$

where $T_k$ denotes the $k$-th decision tree. RF is included because threshold-like conflict behavior may be captured through recursive nonlinear partitioning of the handcrafted surrogate safety feature space.

*Extreme Gradient Boosting (XGBoost)*
XGBoost is a sequential boosting framework introduced by Chen and Guestrin (2016) that iteratively adds new trees to minimize the residual error of the previous ensemble. Its additive prediction function is defined as:

$$\hat{y} = \sum_{k=1}^{K} f_k\,(z^{ML}) \tag{33}$$

where $f_k$ is the $k$-th boosted tree learner. Because XGBoost performs gradient-guided residual optimization, it can learn highly nonlinear interactions among the handcrafted traffic safety descriptors.

*Light Gradient Boosting Machine (LightGBM)*
LightGBM follows a similar additive boosting formulation introduced by Ke et al. (2017):

$$\hat{y} = \sum_{k=1}^{K} g_k\,(z^{ML}) \tag{34}$$

where $g_k$ denotes the $k$-th LightGBM learner. Its histogram-based split finding and leaf-wise growth allow efficient learning of sparse interaction patterns and provide an additional high-performance tree ensemble comparator.

*3.10.1.6. Support Vector Machine (SVM)*
SVM is a margin-maximization classifier that estimates the optimal separating hyperplane in transformed kernel space. Its decision function is:

$$\hat{y} = \text{sign}\left(\sum_{i=1}^{n} \alpha_i\, y_i K(z_i, z^{ML}) + b\right) \tag{35}$$

where $\alpha_i$ are support vector coefficients, $K(\cdot)$ is the kernel similarity function, and $b$ is the intercept. Because SVM can model nonlinear separation under kernel transformation while remaining robust under moderate-dimensional handcrafted feature spaces, it provides an important margin-based baseline.

<u>Deep Learning Tier</u>
The DL tier includes Bidirectional Long Short-Term Memory (BiLSTM), GRU, TCN, and Transformer models. These architectures preserve frame order but do not explicitly represent heterogeneous graph topology. Each frame was summarized independently using pooled node, edge, and graph-level descriptors. For the node set $V_t$, mean and maximum pooling were applied to the seven normalized continuous node features:

$$x_{node,t} = [\mu(V_t) \parallel \max(V_t)] \in \mathbb{R}^{14} \quad (36)$$

For the edge set $E_t$, mean, maximum, and minimum pooling were applied over the six normalized edge descriptors:

$$x_{edge,t} = [\mu(E_t) \parallel \max(E_t) \parallel \min(E_t)] \in \mathbb{R}^{18} \quad (37)$$

Three graph-level scalar variables were appended:

$$x_{graph,t} = [N_t,\ M_t,\ \rho_t^{veh}] \quad (38)$$

Thus, each frame was represented as:

$$x_t^{DL} = \left[x_{node,t} \parallel x_{edge,t} \parallel x_{graph,t}\right] \in \mathbb{R}^{35} \quad (39)$$

and each temporal sample became:

$$X^{DL} = x_1^{DL}, x_2^{DL}, \dots, x_T^{DL} \in \mathbb{R}^{8\times35} \quad (40)$$

which follows the implemented flat sequence dataset generation.

*Bidirectional LSTM (BiLSTM)*

BiLSTM processes the temporal sequence in both forward and backward directions, allowing the model to capture past-future contextual dependence:

$$h_t = \Psi_f(x_t, h_{t-1}) \parallel \Psi_b(x_t, h_{t+1}) \quad (41)$$

where $\Psi_f$and $\Psi_b$denote forward and backward LSTM recurrence.

*Gated Recurrent Unit (GRU)*

GRU provides a lighter recurrent temporal model with gated hidden-state updating:

$$h_t = \Psi_{GRU}(x_t, h_{t-1}) \quad (42)$$

which retains temporal memory while reducing recurrent parameter complexity.

*Temporal Convolutional Network (TCN)*

TCN applies one-dimensional dilated temporal convolution over the frame sequence:

$$H = \mathrm{Conv1D}(X^{DL}) \quad (43)$$

allowing short-range and medium-range temporal conflict signatures to be extracted without recurrent hidden-state propagation.

*Transformer*

The Transformer applies self-attention across the ordered frame embeddings:

$$H = \text{SelfAttention}(X^{DL}) \tag{44}$$

This allows each frame to attend to all other frames within the observation window. Collectively, the DL tier assesses whether temporal patterns in handcrafted frame descriptors can identify conflict escalation without explicit relational graph modeling.

Graph Tier

The graph tier processes the complete heterogeneous scene-graph sequence without handcrafted flattening, preserving node attributes, relation-specific adjacency structures, and pairwise surrogate-safety edge features. This comparison evaluates whether deeper, co-evolving relational graph modeling provides predictive gains over a shallower graph baseline and non-graph temporal models.

*Baseline HERMES*

As the foundational graph benchmark, the baseline model is structurally restricted to a single-stage, relation-specific, SSM-aware graph propagation layer ($\ell = 1$) followed by DRAC-guided graph pooling and temporal LSTM aggregation. Unlike the multi-head scaled dot-product mechanics of the proposed model, the baseline utilizes a single-head additive spatial attention mechanism.

For each relation type $r \in \{\text{VV}, \text{VP}, \text{PP}\}$, the spatial hidden representation of node $i$ is derived through a pairwise concatenation of projected source and target node states, supplemented by an edge-derived structural attention bias:

$$h_{t,i}^{r} = W_r X_{t,i} + b_r \tag{45}$$

The encoded edge attributes are averaged across channels to obtain a scalar structural bias:

$$B_{t,ij} = \left[ \frac{1}{F_e} \sum_{c=1}^{F_e} \left( E_{t,ij} W_e + b_e \right)_c \right] \tag{46}$$

where $W_r \in \mathbb{R}^{F_n \times D_{\text{hidden}}}$ projects the raw attributes, and $B_{t,ij}$ averages the channel projections of the encoded edge attributes to yield a scalar structural bias.

The unnormalized spatial attention scores are mapped through a single-layer feed-forward network combined with a LeakyReLU non-linearity:

$$S_{t,ij}^{r} = \text{LeakyReLU}\left( W_{\text{attn}}^{r} \left[ h_{t,i}^{r} \| h_{t,j}^{r} \right] + b_{\text{attn}}^{r} \right) + B_{t,ij} \tag{47}$$

Topological boundaries are enforced by masking unconnected node combinations prior to running a softmax normalization over incoming neighbors:

$$\tilde{S}_{t,ij}^{r} = \begin{cases} S_{t,ij}^{r}, & A_{t,ij}^{r} = 1 \\ -\infty, & A_{t,ij}^{r} = 0 \end{cases} \tag{48}$$

The normalized attention coefficient is:

$$\alpha_{t,ij}^{r} = \text{softmax}_j\left(\tilde{S}_{t,ij}^{r}\right) \tag{49}$$

The relation-specific incoming message vector $M_t^r$ aggregates neighbor representations via a batch matrix multiplication. The relation-specific message received by node $i$ is:

$$M_{t,i}^{r} = \sum_{j \in \mathcal{N}_i^r} \alpha_{t,ij}^{r}\, h_{t,j}^{r} \tag{50}$$

The independent spatial messages are fused directly with the original input attributes $X_t$, passing through layer normalization to compute the baseline node state matrix $H_t \in \mathbb{R}^{N \times D_{\text{hidden}}}$:

$$H_t = \text{LayerNorm}\left(\text{ReLU}\left([X_t \| M_t^{\text{VV}} \| M_t^{\text{VP}} \| M_t^{\text{PP}}] W_{\text{fusion}} + b_{\text{fusion}}\right)\right) \tag{51}$$

Because Baseline HERMES does not include dynamic edge refinement, DRAC-guided pooling uses the original edge tensor $\text{E}_t$. The node-level risk score is defined as:

$$s_{t,i} = \max_j E_{t,ij,k_{\text{DRAC}}} \tag{52}$$

and the pooled frame representation is

$$g_t = \sum_{i=1}^{N} \left( \frac{\exp\left(s_{t,i}\right)}{\sum_{k=1}^{N} \exp\left(s_{t,k}\right)} \right) H_{t,i} \tag{53}$$

The chronological sequence of pooled frame graph embeddings $H_{\text{seq}} = \text{Stack}(g_1, \dots, g_T) \in \mathbb{R}^{T \times D_{\text{hidden}}}$ updates the recurrent temporal modeling block:

$$Z, _ = \text{LSTM}_{\text{baseline}}\left(H_{\text{seq}}\right) \tag{54}$$

where the terminal output state $z_T = Z[:, T, :]$ is passed through a fully connected layer to compute the binary scene-level conflict logit $\hat{y}$. The baseline architecture is illustrated in Figure 3.

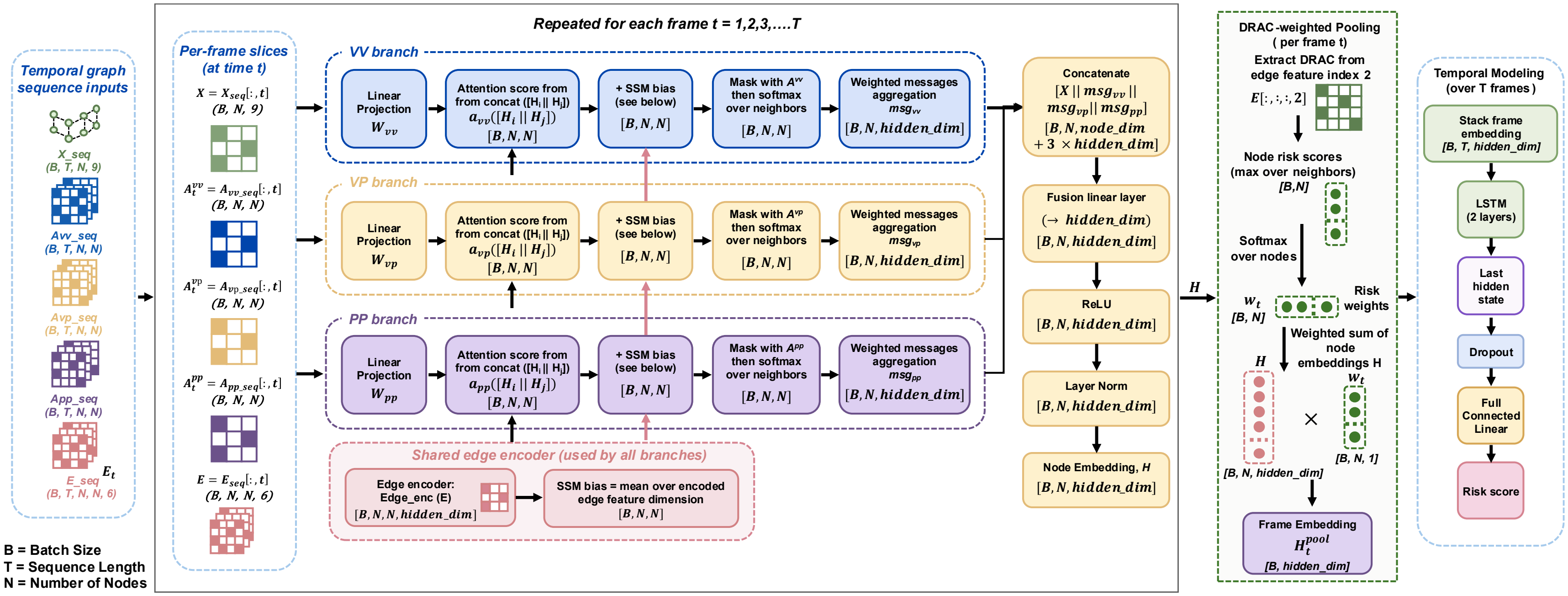

Figure 3: Baseline HERMES architecture.

*Proposed HERMES*

While the baseline graph model preserves heterogeneous interaction topologies and surrogate-safety-aware relational aggregation, it relies on a shallow, single-stage graph propagation process with static edge descriptors. Consequently, relational interactions are encoded only once per frame, and no adaptive feedback occurs between evolving node embeddings and edge safety representations.

To address this structural limitation, the proposed HERMES architecture introduces four interconnected graph-tier enhancements:

- **Dual-Layer Multi-Head Graph Propagation:** Expands spatial capacity by stacking two deep sequential propagation blocks ($\ell = 2$) using scaled dot-product attention split over multiple parallel head channels to monitor complementary risk interpretations.
- **Dynamic Edge Refinement:** Interleaves an explicit node-to-edge update block ($E_t^{(\ell)}$) between successive layers, allowing latent node states to continuously update interaction semantics.
- **Deeper DRAC-Prioritized Graph Summarization:** Leverages the dynamically refined edge features from the terminal graph layer ($E_t^{(2)}$) to compute highly responsive, co-evolved pooling weights.
- **Fully Integrated Temporal Safety Optimization:** Backpropagates gradients jointly through the synchronous structural updates and temporal transitions to minimize overall sequence classification errors under an end-to-end learning objective.

Accordingly, Baseline HERMES follows a static edge-representation pathway:

$$\mathcal{F}_{\text{baseline}}: \mathcal{S} \longrightarrow \{H_t, E_t\}_{t=1}^{T} \longrightarrow g_t \longrightarrow z_T \longrightarrow \hat{y} \tag{55}$$

whereas Enhanced HERMES learns jointly refined node and edge representations:

$$\mathcal{F}_{\text{proposed}}: \mathcal{S} \longrightarrow \left\{H_t^{(2)}, E_t^{(2)}\right\}_{t=1}^{T} \longrightarrow g_t \longrightarrow z_T \longrightarrow \hat{y} \tag{56}$$

through iterative node-edge relational interaction and safety-aware temporal representation learning.

Thus, the direct comparison between the two architectural frameworks provides an explicit graph-tier ablation setup: the baseline evaluates the predictive value of generic heterogeneous graph sequence learning, whereas the proposed framework evaluates whether dynamic, surrogate-safety-aware co-evolutionary refinements yield measurable conflict prediction gains under real-world traffic constraints.

## 3.3 Training Configuration

<u>Loss Function and Class-Imbalance Handling</u>

All neural network models were trained using binary cross-entropy with logits loss, implemented through `BCEWithLogitsLoss`. This formulation combines the sigmoid transformation and binary cross-entropy calculation within a numerically stable operation.

To account for the imbalance between safe and conflict sequences in the San Marcos training partition, the positive-class weight was defined as:

$$w_{+} = \frac{N_0}{N_1} \tag{57}$$

where $N_0$ and $N_1$ denote the numbers of safe and conflict sequences, respectively. The positive-class weight increased the contribution of incorrectly classified conflict sequences to the training loss, thereby reducing the tendency of the model to favor the majority safe class.

For conventional machine-learning models that supported class weighting, inverse-frequency weighting was applied through the $\texttt{class_weight} = \texttt{'balanced'}$ option. For XGBoost, the equivalent imbalance adjustment was implemented using

$$\texttt{scale_pos_weight} = \frac{N_0}{N_1} \tag{58}$$

The class weights were calculated exclusively from the training partition and were not recomputed using validation or test labels.

Optimizer and Regularization

All neural network models were optimized using Adam with an initial learning rate of $10^{-3}$ and an $L_2$ weight-decay coefficient of $10^{-4}$. Adam was selected because its adaptive parameter-specific updates provide stable convergence across models containing recurrent, attention-based, and graph message-passing components.

Learning-Rate Scheduling

A `ReduceLROnPlateau` scheduler monitored validation Area Under the Receiver Operating Characteristic Curve (AUC-ROC) after each training epoch. The scheduler operated in maximization mode and reduced the learning rate by a factor of 0.5 when validation AUC-ROC failed to improve for three consecutive epochs. This strategy allowed larger parameter updates during the early stages of optimization and progressively smaller updates as validation performance approached a plateau.

Gradient Clipping

Before each parameter update, the global $L_2$ norm of the gradients was clipped to a maximum value of 1.0. Gradient clipping was applied to stabilize the optimization of the recurrent and graph-based architectures and to limit occasional high-magnitude updates resulting from temporal recurrence or graph message aggregation.

Early Stopping and Model Selection

Graph-based models were trained for a maximum of 50 epochs, whereas deep-learning baseline models were trained for a maximum of 100 epochs. Validation AUC-ROC was used as the model-selection criterion. A model checkpoint was saved whenever validation AUC-ROC improved, and training was terminated when no improvement was observed for five consecutive epochs. The checkpoint associated with the highest validation AUC-ROC was restored before final evaluation on the test partition.

Batch Size

A batch size of 16 was used for all graph and deep-learning models. For the graph architectures, this value was selected from the candidate set $\{8,16,32,64\}$ based on GPU-memory requirements and preliminary optimization behavior. The same batch size was retained for the deep-learning baselines to maintain a consistent training configuration across neural model families.

Multi-Seed Evaluation
All stochastic neural models in the deep-learning and graph tiers were trained using five independent random seeds: {42,123,456,789,2024}. For each run, the random states of Python, NumPy, PyTorch CPU operations, and PyTorch CUDA operations were initialized using the corresponding seed. cuDNN deterministic mode was enabled and benchmarking was disabled to reduce nondeterministic variation. Model parameters were reinitialized independently for each run. Test performance for the stochastic neural models was reported as the mean and standard deviation across the five seeds. This repeated-training design quantified variability associated with random parameter initialization and stochastic optimization.

Machine-Learning Hyperparameter Selection
The machine-learning models were tuned using grid search with AUC-ROC as the optimization criterion. Hyperparameter selection was conducted using only the San Marcos training partition. To prevent temporally overlapping sequences from the same video segment from appearing in both fitting and validation folds, cross-validation folds were defined at the video-segment level. After hyperparameter selection, each machine-learning model was refitted using the complete training partition and evaluated on the held-out test partition. Where applicable, model-specific random states were fixed to ensure reproducibility.

Computational Environment
Neural network training was conducted on a workstation equipped with an Intel Xeon Silver 4214R processor with 12 physical cores and 24 logical processors, 64 GB of system memory, and an NVIDIA RTX A4000 GPU with 16 GB of video memory. Data loading used a single worker process (`num_workers = 0`). Mixed-precision training was not used.

### 3.4 Hyperparameter Configuration

Hyperparameters were selected separately for the ML, DL, and graph tiers because of their distinct architectures and optimization requirements, with validation AUC-ROC used as the primary selection criterion. ML configurations were evaluated through three-fold stratified grid search on the San Marcos training descriptors. DL hyperparameters, including hidden dimension, depth, kernel size, attention-head count, and dropout, were examined using the San Marcos validation set. Baseline and Enhanced HERMES used the same node and edge inputs, hidden dimensions, LSTM depth, and dropout, differing only in the graph-processing components under evaluation. Table 2 summarizes the final benchmark configurations.

Table 2: Final hyperparameter configurations used for benchmark evaluation

| **Model tier** | **Model** | **Final hyperparameters** |
|---|---|---|
| **Machine Learning** | Logistic Regression | C = 0.01, penalty = L2, solver = saga, class_weight = balanced |
| | Random Forest | n_estimators = 500, max_depth = None, min_samples_leaf = 1, class_weight = balanced |
| | XGBoost | n_estimators = 500, max_depth = 7, learning_rate = 0.1, subsample = 0.9, scale_pos_weight = imbalance-adjusted |
| | LightGBM | n_estimators = 500, max_depth = 20, num_leaves = 127, learning_rate = 0.1, class_weight = balanced |
| | SVM | C = 10.0, gamma = scale, kernel = RBF, class_weight = balanced |
| **Deep Learning** | BiLSTM | input_dim = 35, hidden_dim = 128, num_layers = 3, dropout = 0.30, batch_size = 16 |
| | GRU | input_dim = 35, hidden_dim = 64, num_layers = 2, dropout = 0.30, batch_size = 16 |
| | TCN | input_dim = 35, hidden_dim = 64, num_layers = 3, kernel_size = 3, dropout = 0.30, batch_size = 16 |
| | Transformer | input_dim = 35, hidden_dim = 32, num_layers = 3, num_heads = 4, max_seq_len = 8, dropout = 0.30, batch_size = 16 |
| **Graph** | Baseline HERMES | node_dim = 9, edge_dim = 6, graph_hidden = 64, graph_layers = 1, lstm_hidden = 64, lstm_layers = 2, dropout = 0.30, batch_size = 16, params = 82,436 |
| | Enhanced HERMES | node_dim = 9, edge_dim = 6, graph_hidden = 64, graph_layers = 2, num_heads = 4, lstm_hidden = 64, lstm_layers = 2, dropout = 0.30, batch_size = 16, params = 202,069 |

### 3.5 Evaluation Protocol

All models were evaluated for discrimination, threshold-dependent classification, probability quality, low-false-alarm performance, and run-to-run stability. The protocol comprised evaluation on the held-out San Marcos test partition, repeated multi-seed assessment of stochastic neural models, and transfer analysis using the independent Coldwater dataset.

The San Marcos validation partition was used for hyperparameter selection, early stopping, and checkpoint selection, while the San Marcos test partition was reserved for final internal evaluation. Cross-site performance was assessed in two settings. First, the San Marcos-trained model was evaluated directly on the fixed Coldwater test partition without target-site training. Second, mixed-source models were trained from random initialization using all San Marcos training sequences together with selected Coldwater training segments, validated on the fixed Coldwater validation partition, and evaluated on the unchanged Coldwater test partition.

For threshold-dependent metrics, predictions were classified using $\tau = 0.5$. Sensitivity at fixed false-alarm rates was computed from the receiver operating characteristic curve. Let $TP$, $TN$, $FP$, and $FN$ denote the numbers of true positives, true negatives, false positives, and false negatives, respectively.

### 3.6 Performance Metrics

Because the San Marcos dataset exhibited a safe-to-conflict imbalance ratio of approximately $4.3{:}1$, accuracy alone was insufficient for model evaluation. A classifier predicting every sequence as safe would achieve approximately $81.1\%$ accuracy while detecting no conflicts. Performance was therefore evaluated using complementary threshold-dependent, threshold-independent, agreement, calibration, and operational metrics.

Threshold-Dependent Classification Metrics

Accuracy was computed as:

$$\text{Accuracy} = \frac{TP + TN}{TP + TN + FP + FN} \tag{59}$$

Precision measured the proportion of predicted conflicts that corresponded to actual conflict sequences:

$$\text{Precision} = \frac{TP}{TP + FP} \tag{60}$$

Recall, also referred to as sensitivity or the true-positive rate, measured the proportion of actual conflicts correctly identified:

$$\text{Recall} = \frac{TP}{TP + FN} \tag{61}$$

Specificity measured the proportion of safe sequences correctly classified:

$$\text{Specificity} = \frac{TN}{TN + FP} \tag{62}$$

The F1 score was computed as the harmonic-mean of precision and recall:

$$\text{F1} = 2\,\frac{\text{Precision} \times \text{Recall}}{\text{Precision} + \text{Recall}} \tag{63}$$

These metrics characterized the binary classification behavior of each model at the specified operating threshold.

Threshold-Independent Ranking Metrics

AUC-ROC was used to evaluate the ability of each model to rank conflict sequences above safe sequences across all possible classification thresholds. AUC-ROC can be interpreted as the probability that a randomly selected conflict sequence receives a higher predicted score than a randomly selected safe sequence. It served as the primary model-selection and hyperparameter-tuning metric.

The Area Under the Precision-Recall Curve (AUC-PR) was computed using average precision:

$$\text{AUC-PR} = \sum_{n} (R_n - R_{n-1})\, P_n \tag{64}$$

where $P_n$ and $R_n$ denote precision and recall at the $n$-th operating threshold. AUC-PR provides a minority-class-focused measure and is particularly informative when the conflict class is substantially less frequent than the safe class.

Balanced Agreement and Probability Metrics

The Matthews Correlation Coefficient (MCC) was calculated as:

$$\text{MCC} = \frac{TP \times TN - FP \times FN}{\sqrt{(TP + FP)(TP + FN)(TN + FP)(TN + FN)}} \tag{65}$$

MCC incorporates all four confusion-matrix components and provides a balanced measure of binary classification performance under class imbalance.

Cohen's kappa was used to quantify agreement between predicted and observed labels beyond agreement expected by chance:

$$\kappa = \frac{p_o - p_e}{1 - p_e} \tag{66}$$

where $p_o$ is the observed agreement and $p_e$ is the expected agreement under chance-level classification.

The Brier score was computed as:

$$\text{Brier} = \frac{1}{N}\sum_{i=1}^{N}(\hat{p}_i - y_i)^2 \tag{67}$$

where $\hat{p}_i$ is the predicted probability of conflict and $y_i \in \{0,1\}$ is the observed label. Lower Brier scores indicate smaller probabilistic prediction error. Because class-weighted loss functions can alter probability calibration, the Brier score was interpreted as a measure of overall probabilistic accuracy rather than as sufficient evidence of calibration by itself.

Operational False-Alarm-Constrained Sensitivity

For practical warning applications, conflict sensitivity was evaluated under predefined false-alarm constraints. The false-positive rate was defined as:

$$\text{FPR} = \frac{FP}{FP + TN} \tag{68}$$

Sensitivity at an $X\%$ false-alarm rate was defined as the maximum true-positive rate attainable while satisfying $\text{FPR} \leq X/100$:

$$\text{S@X\% FAR} = \max_{\tau:\, \text{FPR}(\tau) \leq X/100} \text{TPR}(\tau) \tag{69}$$

where

$$\text{TPR}(\tau) = \frac{TP(\tau)}{TP(\tau) + FN(\tau)} \tag{70}$$

Performance was reported at $5\%$ and $10\%$ false-alarm rates, with $\text{S@5\%FAR}$ treated as the principal low-false-alarm metric. For each model, these values were obtained directly from the ROC curve on the held-out test partition. Final test metrics were computed using the checkpoint with the highest validation AUC-ROC, and stochastic neural models were summarized across five independent seeds.

### 3.7 Multi-Seed Statistical Evaluation

To quantify run-to-run variability, all stochastic DL and graph models were trained using five seeds, $\{42,123,456,789,2024\}$. For each run, Python, NumPy, PyTorch CPU, and PyTorch CUDA random states were initialized with the corresponding seed. cuDNN deterministic mode was enabled, benchmark-based kernel selection was disabled, and model parameters were reinitialized before training. All runs used the same training, validation, and test partitions.

The checkpoint with the highest validation AUC-ROC was restored for test evaluation. For each metric, performance was summarized as:

$$\bar{x} = \frac{1}{K}\sum_{k=1}^{K} x_k, \qquad s = \sqrt{\frac{1}{K-1}\sum_{k=1}^{K}(x_k - \bar{x})^2} \tag{71}$$

where $\bar{x}$ and $s$ denote the mean and sample standard deviation across $K = 5$ runs. The mean represents average predictive performance, while the standard deviation reflects sensitivity to stochastic initialization and training order.

Seed-matched AUC-ROC differences between HERMES and each stochastic neural baseline were also examined. Because identical seeds were used across models, each HERMES run was paired with the corresponding baseline run, and the paired differences were summarized by their direction, magnitude, mean, and standard deviation.

The Wilcoxon signed-rank test was used as an exploratory nonparametric comparison of the five paired AUC-ROC values. With only five nonzero pairs, the smallest attainable two-sided exact $p$-value is $0.0625$. The test was therefore interpreted as descriptive evidence of cross-seed consistency rather than as a definitive test of statistical significance at $\alpha = 0.05$.

Conventional ML models were not retrained across the same five seeds. Stochastic ML algorithms used fixed random states for reproducibility, and their variability was assessed through cross-validation during hyperparameter selection.

### 3.8 Ablation Study Design

A structured ablation study was conducted to quantify the contribution of the principal design components of HERMES. The study was organized into four experimental groups corresponding to edge-feature richness, graph architecture, temporal encoding, and heterogeneous interaction-type coverage. Within each group, variants were defined progressively so that one architectural dimension was modified while the remaining model configuration, optimizer settings, data partitions, and evaluation procedure were held constant. All ablation variants were trained from scratch using the same five random seeds, and their performance was summarized using the mean and standard deviation of the test metrics. The complete experimental scheme is presented in Table 3.

Table 3: Experimental configurations used in the HERMES ablation study.

| Level | A - SSM Edge Features | B - Graph Architecture | C - Temporal Encoder | D - Interaction Types |
|---|---|---|---|---|
| **0** | A0: Nodes only, no edges | | | |
| **1** | A1: + kinematics (distance, rel_speed) | B1: No graph, mean pool only | C1: Single frame, no temporal | D1: VV edges only |
| **2** | A2: + DRAC | B2: Simple GNN, uniform adjacency | C2: GNN + GRU | D2: VP edges only |
| **3** | A3: + heading_diff | B3: SSM attention + mean pooling | C3: GNN + Transformer | D3: VV + VP only |
| **4** | A4: Full 6-dim edges | B4: Full baseline HERMES | C4: GNN + LSTM | D4: All types (VV+VP+PP) |

Group A: SSM Edge-Feature Contribution

Group A examined the incremental value of relational edge information beyond node-level kinematic attributes. The complete HERMES edge vector was defined as:

$$\mathbf{e}_{ij} = \left[d_{ij}, v_{\mathrm{rel},ij}, \mathrm{DRAC}_{ij}, \Delta\theta_{ij}, \Delta v_{x,ij}, \Delta v_{y,ij}\right]^{\top} \tag{72}$$

where $d_{ij}$ denotes inter-agent distance, $v_{\mathrm{rel},ij}$ is the magnitude of relative velocity, $\mathrm{DRAC}_{ij}$ represents braking demand, $\Delta\theta_{ij}$ is the absolute heading difference, and $\Delta v_{x,ij}$ and $\Delta v_{y,ij}$ are the relative velocity components. Variant A0 removed all interaction edges and relied only on node features, whereas A1 introduced basic kinematic relations through distance and relative speed. A2 added DRAC to determine whether a physics-informed braking-demand descriptor provided predictive information beyond proximity and relative motion. A3 further included heading difference to account for directional alignment, and A4 used the full six-dimensional edge representation. This progression evaluated whether increasingly detailed interaction physics improved the model's ability to distinguish safety-critical scene configurations.

Group B: Graph-Architecture Contribution

Group B isolated the contribution of graph-based relational modeling and the principal graph operations introduced in HERMES. Variant B1 removed graph message passing entirely; node features were projected and averaged within each frame before temporal modeling. B2 introduced a simple GNN using uniform adjacency-based propagation, thereby testing whether relational connectivity alone improved prediction. B3 retained the SSM-informed attention mechanism but replaced DRAC-guided graph pooling with ordinary mean pooling, allowing the effect of safety-aware message weighting to be separated from risk-sensitive graph summarization. B4 represented the complete HERMES graph architecture, including relation-specific attention and DRAC-guided pooling. The architectural progression can be expressed as:

$$\begin{aligned}&\text{Node aggregation} \rightarrow \text{Generic message passing} \rightarrow \text{SSM-informed attention}\\ &\rightarrow \text{DRAC-guided graph pooling}\end{aligned} \tag{73}$$

This group therefore assessed whether performance gains arose from graph connectivity alone, from the use of SSM-biased attention, or from prioritizing agents associated with greater braking demand during frame-level representation learning.

Group C: Temporal-Encoder Contribution

Group C evaluated the contribution of short-term temporal modeling to conflict prediction. Let $\mathbf{g}_t$ denote the graph-level scene representation at frame $t$. The generic temporal prediction task was formulated as:

$$\hat{y} = f_{\mathrm{temp}}(\mathbf{g}_1, \mathbf{g}_2, \dots, \mathbf{g}_T) \tag{74}$$

where $f_{\mathrm{temp}}$ denotes the temporal aggregation mechanism. Variant C1 used only a single frame-level graph embedding and therefore contained no temporal encoder. C2 applied a GRU to the ordered graph sequence, C3 used a Transformer encoder with self-attention and positional embeddings, and C4 used the LSTM configuration adopted in the full HERMES framework. This comparison determined whether conflict prediction was driven primarily by instantaneous graph structure or by the short-term evolution of interaction risk across the eight-frame observation window. It also evaluated whether recurrent or attention-based temporal encoders were better suited to the observed duration and structure of the graph sequences.

Group D: Interaction-Type Contribution
Group D examined the contribution of heterogeneous relation coverage. The complete frame-level graph contained three relation-specific adjacency channels:

$$\mathcal{A}_t = \{\mathbf{A}_t^{\mathrm{VV}}, \mathbf{A}_t^{\mathrm{VP}}, \mathbf{A}_t^{\mathrm{PP}}\} \tag{75}$$

where $\mathbf{A}_t^{\mathrm{VV}}$, $\mathbf{A}_t^{\mathrm{VP}}$, and $\mathbf{A}_t^{\mathrm{PP}}$ represent vehicle-vehicle, vehicle-pedestrian, and pedestrian-pedestrian interactions, respectively. Variant D1 retained only VV edges, whereas D2 retained only VP edges. D3 combined VV and VP relations, which directly correspond to the interaction types considered in vehicle-related conflict labeling. D4 retained all three relation channels, including PP edges, to determine whether pedestrian-group context contributed useful scene-level information even though PP interactions were not assigned conflict labels. This group therefore assessed the relative importance of direct vehicle-related interactions and broader contextual relations within the heterogeneous traffic scene.

Ablation Evaluation Criteria
The four ablation groups collectively evaluated the HERMES framework along complementary dimensions of interaction representation, graph reasoning, temporal learning, and relational coverage. Because only one design dimension was modified within each group, differences in predictive performance could be associated with the corresponding architectural change while minimizing confounding from unrelated implementation choices. The ablation analysis was therefore used to determine whether the final HERMES configuration benefited consistently from richer edge information, safety-informed graph operations, temporal sequence modeling, and complete heterogeneous interaction coverage.

### 3.9 Cross-Site Transferability Evaluation

Cross-site performance was assessed using San Marcos as the source site and Coldwater, Michigan, as an independent but broadly comparable target site. Two settings were evaluated: zero-shot transfer and mixed-source training.

In the zero-shot setting, the HERMES model trained on the complete San Marcos training partition was applied directly to the fixed Coldwater test set without exposure to Coldwater training data. This experiment assessed whether the learned spatial, kinematic, and relational representations transferred to an unseen but similar signalized-intersection environment.

Mixed-source training was then conducted by combining all San Marcos training sequences with subsets of the Coldwater training data. The evaluated target-site segment ratios were

$$\mathcal{R} = \{10\%, 20\%, 30\%, 40\%, 50\%, 60\%\} \tag{76}$$

For each ratio $r$, the training set was defined as:

$$\mathcal{D}_{\mathrm{train}}^{(r)} = \mathcal{D}_{\mathrm{SM}}^{\mathrm{train}} \cup \mathcal{D}_{\mathrm{CW}}^{(r)} \tag{77}$$

where $\mathcal{D}_{\mathrm{SM}}^{\mathrm{train}}$ contains all San Marcos training sequences and $\mathcal{D}_{\mathrm{CW}}^{(r)}$ contains sequences from a randomly selected proportion of the 93 Coldwater training video segments. Because segment lengths differed, the number of included Coldwater sequences varied across seeds at the same nominal ratio.

For each ratio, a new HERMES model was trained from random initialization using three seeds. Model selection was based on the fixed Coldwater validation partition, and final performance was evaluated on the unchanged Coldwater test partition. The test data were excluded from model fitting and checkpoint selection.

This design distinguishes direct zero-shot transfer from joint source-target training and isolates the effect of adding limited target-site information while retaining the complete source-site training set.

# 4 DATA

## 4.1 Study Sites

Trajectory data were collected from two geographically independent but operationally comparable signalized intersections. The San Marcos, Texas site was used for model development, validation, hyperparameter tuning, and internal evaluation. An external dataset from Coldwater, Michigan was used to assess out-of-sample transferability under broadly similar lane configurations, signal control, and pedestrian interaction conditions. Using comparable sites reduces major geometric and operational confounding while testing whether the learned spatiotemporal interaction patterns generalize across independently collected urban intersection data. Table 4 summarizes the geographic, geometric, and operational characteristics of both sites.

Table 4: Summary of the San Marcos benchmark and Coldwater transferability datasets.

| **Intersection** | **San Marcos, Texas** | **Coldwater, Michigan** |
|---|---|---|
| Coordinates | 29.826361°N, 97.983844°W | 41.940622°N, 85.000745°W |
| Camera View | | |
| Satellite View | | |
| Temporal Window | August 9, 2025; 9h 47m (9:51 AM to 7:39 PM) | September 20, 202; 12h 15m (06:33 AM to 06:49 PM) |
| Land Use Profile | Urban commercial corridor | Urban commercial corridor |
| Approach Layout | 4-leg, multi-lane signalized | 4-leg, multi-lane signalized |
| Crossing Infrastructure | Marked crosswalks / brick curb nodes | Marked crosswalks / brick curb nodes |
| Methodological Assignment | Base training, validation, testing | Zero-shot evaluation, domain adaptation |
| Frame graphs | 505,061 | 266,901 |
| Total Sequences | 109,028 | 47,367 |
| Conflict sequences | 20,576 (18.9%) | 4,536 (9.6%) |

## 4.2 Data Processing

Trajectory Processing

Video data from both sites were processed through a unified detection and tracking pipeline. Road users were detected frame by frame using YOLOv8n at the native video resolution, with a confidence threshold of 0.25 and an NMS IoU threshold of 0.40. ByteTrack was then used to preserve identities across frames by associating high-confidence detections first and recovering lower-confidence detections affected by occlusion or scale variation.

The resulting records contained bounding-box coordinates, timestamps, and confidence scores. Trajectories were uniquely identified using the combination of `video_id` and `track_id`, and duplicate records were removed before geometric processing. The pipeline generated approximately 13.5 million trajectory records for San Marcos and 14.5 million for Coldwater. Six detected classes were grouped into two operational categories: vehicles, including cars, trucks, buses, motorcycles, and bicycles, and pedestrians. The extracted trajectories were subsequently smoothed using Bézier curve fitting and Savitzky-Golay filtering.

A projective homography was used to transform image coordinates into georeferenced ground-plane coordinates. The bottom center of each bounding box was treated as the road-user ground contact point,

$$x_{\text{anchor}} = u, \qquad y_{\text{anchor}} = v + \frac{h}{2} \tag{78}$$

A quadrilateral region of interest was defined for each site to retain the principal conflict area and exclude peripheral roadway segments. Points outside the region were removed using a point-in-polygon procedure.

Let $\mathbf{x} = (u, v, 1)^T$ denote an image-plane point and $\mathbf{X} = (X, Y, w)^T$ its corresponding ground-plane coordinate. The homography $\mathbf{H} \in \mathbb{R}^{3\times3}$ satisfies

$$\begin{pmatrix} X \\ Y \\ w \end{pmatrix} \sim \mathbf{H} \begin{pmatrix} u \\ v \\ 1 \end{pmatrix} \tag{79}$$

where $\sim$ denotes projective equivalence up to a nonzero scale factor. Cartesian coordinates were obtained as $X/w$ and $Y/w$. For each site, $\mathbf{H}$ was estimated using singular value decomposition from matched image and georeferenced control points.

The projected geographic coordinates were converted to local metric coordinates relative to a site-specific origin $(lat_0, lon_0)$:

$$x_{\text{local}} = (lon - lon_0)\frac{\pi}{180} R \cos(lat_0) \tag{80}$$

$$y_{\text{local}} = (lat - lat_0)\frac{\pi}{180} R \tag{81}$$

where $R \approx 6{,}378{,}137$ m. A secondary affine normalization then mapped trajectories into a common local reference frame to improve spatial consistency across the two sites.

Trajectory Quality Control, and Kinematic Feature Extraction

Raw trajectories may contain short tracks, temporal gaps, identity switches, and localization jitter that propagate into velocity, acceleration, graph construction, and conflict labels. A sequential quality-control procedure was therefore applied to enforce minimum duration, temporal continuity, spatial consistency, and physical plausibility.

Tracks with fewer than five observations or a maximum internal frame gap exceeding five frames were removed. The five-observation minimum ensured sufficient support for the third-order polynomial used during Savitzky-Golay smoothing. Localization anomalies were identified from the frame-to-frame displacement

$$\Delta d_t = \sqrt{(x_t - x_{t-1})^2 + (y_t - y_{t-1})^2} \tag{82}$$

Observations with $\Delta d_t > 5.0$ m were discarded. At 30 frames per second, this threshold corresponds to approximately $150$ m/s over one frame and was used to identify gross localization errors rather than impose an operational speed limit. The first observation of each trajectory was retained because no preceding point was available. After anomaly removal, track continuity was reassessed.

To accommodate retained temporal gaps, elapsed time was calculated from the observed frame indices:

$$\Delta t_t = \frac{f_t - f_{t-1}}{30} \tag{83}$$

where $f_t$ and $f_{t-1}$ denote successive valid frame indices. This formulation prevents overestimation of derivatives when consecutive retained observations are separated by more than one frame.

The local metric coordinates were smoothed independently using a Savitzky-Golay filter with an 11-frame window and third-order polynomial:

$$p(\tau) = a_0 + a_1\tau + a_2\tau^2 + \cdots + a_n\tau^n \tag{84}$$

where $n = 3$. For trajectories shorter than 11 observations, the largest available odd window exceeding the polynomial order was used, with a minimum admissible window of five observations. Velocity and acceleration were obtained from the first and second analytical derivatives of the fitted coordinates:

$$v_x = \frac{dx}{dt}; v_y = \frac{dy}{dt} \tag{85}$$

$$a_x = \frac{d^2x}{dt^2}; a_y = \frac{d^2y}{dt^2} \tag{86}$$

Scalar speed and acceleration magnitude were then calculated as:

$$speed = \sqrt{{v_x}^2 + {v_y}^2} \tag{87}$$

$$acceleration = \sqrt{{a_x}^2 + {a_y}^2} \tag{88}$$

Analytical differentiation of the local polynomial reduces the noise amplification associated with direct finite differences. Residual kinematic outliers were removed using agent-specific plausibility limits. Vehicle observations were excluded when $v > 20$ m/s or $a > 9.0$ m/s$^2$, thresholds selected to retain plausible high-speed and evasive maneuvers while removing artifacts caused by unstable bounding boxes or identity switches (Formosa et al., 2020; Lyu et al., 2025). Pedestrian observations were excluded when $v > 3.0$ m/s or $a > 4.0$ m/s$^2$, preserving rapid pedestrian responses while filtering implausible motion (Muduli et al., 2026; Zhang et al., 2020b, 2020a).

Derivative estimates unavailable near trajectory boundaries were masked and excluded from plausibility screening and graph-node feature construction, while the corresponding positional observations were retained. After quality control, the San Marcos data decreased from approximately 13.5 million raw observations to 1.3 million validated observations, corresponding to a retention rate of 9.6%. The Coldwater data decreased from approximately 14.5 million to 2.2 million observations, corresponding to a retention rate of 15.2%. The resulting trajectories were used for heterogeneous graph construction and conflict classification.

Surrogate Safety Measure Computation

TTC and DRAC were computed from the filtered trajectories to characterize pairwise collision imminence and braking demand. At each frame, an interaction edge was created between road users $i$ and $j$ when their Euclidean separation was within the predefined interaction radius. Let $\Delta \mathrm{r}_{ij} = \mathrm{r}_j - \mathrm{r}_i, \Delta \mathrm{v}_{ij} = \mathrm{v}_j - \mathrm{v}_i$, where $\mathbf{r}$ and $\mathbf{v}$ denote position and velocity. Pairwise distance and relative speed were defined as:

$$d_{ij} = \left\| \Delta \mathbf{r}_{ij} \right\| \tag{89}$$

$$v_{\mathrm{rel},ij} = \left\| \Delta \mathbf{v}_{ij} \right\| \tag{90}$$

For interactions with a positive relative closing speed, TTC was computed as:

$$\mathrm{TTC}_{ij} = \frac{d_{ij}}{v_{\mathrm{rel},ij}} \tag{91}$$

TTC estimates the remaining time to a potential collision under unchanged motion conditions (Hayward, 1972). Smaller values indicate greater temporal criticality, with values near 1.5 to 2.0 s often treated as critical and values above 3.0 s generally regarded as lower risk (Kumar and Mudgal, 2025). Because this formulation assumes unchanged motion, it is most applicable to converging and car-following interactions and may not fully represent turning, crossing, or evasive maneuvers (Bhattarai et al., 2023).

DRAC was calculated as:

$$\mathrm{DRAC}_{ij} = \frac{v_{\mathrm{rel},ij}^2}{2d_{ij}} \tag{92}$$

DRAC represents the constant deceleration required to eliminate the closing speed before the available separation is exhausted, assuming unchanged kinematic conditions (Cunto and Saccomanno, 2008; Qu et al., 2014). Higher values indicate greater braking demand and interaction severity, while lower values generally correspond to ordinary spacing conditions (Archer, 2005; Guido et al., 2011).

For numerical stability, TTC and DRAC were treated as undefined when $d_{ij} \leq 10^{-3}$ m or $v_{\mathrm{rel},ij} \leq 10^{-3}$ m/s. DRAC values above 10 m/s$^2$ were excluded because they were more likely to reflect residual trajectory noise than plausible interaction dynamics under typical roadway conditions (Chen et al., 2024; Formosa et al., 2020; Muduli and Ghosh, 2024). Although both measures were computed, only DRAC was included in the final edge representation and risk-weighted pooling mechanism.

Conflict Labeling

Agent pairs were first classified as vehicle-vehicle (VV), vehicle-pedestrian (VP), or pedestrian-pedestrian (PP). Interaction-specific distance thresholds, relative-speed gates, and directional filters were then applied to remove physically implausible or non-threatening observations. PP pairs were excluded from conflict labeling because the prediction target was vehicle-related conflict, but they were retained as contextual graph edges.

VV pairs with $\|\Delta \mathbf{v}_{ij}\| < 1.0$ m/s and VP pairs with $\|\Delta \mathbf{v}_{ij}\| < 0.5$ m/s were treated as safe and excluded from further conflict screening. In addition, VV observations were removed when $|\Delta\theta_{ij}| > 120°$ and $d_{ij} > 5.0$ m, where $\Delta\theta_{ij}$ denotes the absolute heading difference and $d_{ij}$ the instantaneous separation. This condition removed spatially separated, opposing-direction vehicles that could exhibit large relative velocities without forming a plausible conflict. Figure 4 summarizes the labeling procedure.

For each qualifying pair, TTC and distance were aggregated over the frames in which both agents were observed: $\mathrm{minTTC}_{ij} = \min_{t \in \mathcal{T}_{ij}} \mathrm{TTC}_{ij}^{t}$, $\mathrm{minDist}_{ij} = \min_{t \in \mathcal{T}_{ij}} d_{ij}^{t}$, where $\mathcal{T}_{ij}$ is the set of co-observed frames. Pair-level severity was assigned using three classes. A VV pair was labeled as a serious conflict when $\mathrm{minTTC}_{ij} < 1.5$ s and $\mathrm{minDist}_{ij} < 3.0$ m, and as a general conflict when $\mathrm{minTTC}_{ij} < 3.0$ s and $\mathrm{minDist}_{ij} < 5.0$ m. A VP pair was labeled as serious when $\mathrm{minTTC}_{ij} < 1.5$ s and $\mathrm{minDist}_{ij} < 2.0$ m, and as general when $\mathrm{minTTC}_{ij} < 3.0$ s and $\mathrm{minDist}_{ij} < 3.0$ m. Serious-conflict criteria took precedence, and all remaining VV and VP pairs were labeled safe.

The 1.5-s TTC threshold represents highly time-critical interactions and is consistent with established serious-conflict criteria (Hydén, 1987), while the 3.0-s threshold captures a broader set of potentially hazardous interactions under unchanged-motion assumptions (Hayward, 1972). To reduce false labeling from spatially separated trajectories, TTC thresholds were combined with interaction-specific distance limits derived from the simplified stopping-distance approximation $d_{\mathrm{stop}} = v t_r + \frac{v t_b}{2}$, where $v$ is the initial closing speed, $t_r$ is perception-reaction time, and $t_b$ is braking duration. With the adopted parameters, $d_{\mathrm{stop}} \approx 1.1v$. The VV thresholds of 5.0 and 3.0 m correspond approximately to closing speeds of 16 and 10 km/h, respectively. The VP thresholds were set to 3.0 m for general conflicts and 2.0 m for serious conflicts. These values were treated as operational labeling parameters rather than empirically calibrated site-specific safety boundaries.

Among the 33,221 qualifying pairs, 84.5% were labeled safe, 9.7% as general conflicts, and 5.8% as serious conflicts. General and serious conflicts were combined for binary prediction:

$$y_{ij}^{\mathrm{binary}} = \begin{cases} 0, & y_{ij}^{\mathrm{severity}} = 0 \\ 1, & y_{ij}^{\mathrm{severity}} \in \{1,2\} \end{cases} \tag{93}$$

Frame $t$ was labeled as a conflict frame when at least one qualifying VV or VP interaction satisfied the instantaneous TTC and distance criteria:

$$y_t = \mathbb{I}\left[\exists (i,j) \in \mathcal{E}_t^{\mathrm{VV \cup VP}} : \mathrm{TTC}_{ij}^t < 3.0\ \mathrm{s} \ \land \ d_{ij}^t < d_{\mathrm{thr}}^{(r_{ij})}\right] \tag{94}$$

where $\mathbb{I}[\cdot]$ is the indicator function, $\mathcal{E}_t^{\mathrm{VV \cup VP}}$ is the set of qualifying vehicle-related interactions at frame $t$, $r_{ij}$ denotes the interaction type, and

$$d_{\mathrm{thr}}^{(r_{ij})} = \begin{cases} 5.0\ \mathrm{m}, & r_{ij} = \mathrm{VV} \\ 3.0\ \mathrm{m}, & r_{ij} = \mathrm{VP} \end{cases} \tag{95}$$

A temporal sequence containing $T = 8$ consecutive frames was labeled as a conflict sequence when at least one constituent frame was labeled as a conflict:

$$y_{\mathrm{seq}} = \max_{t \in \{1,\ldots,T\}} y_t \tag{96}$$

The final datasets contained 109,028 San Marcos sequences with an 18.9% conflict rate and 47,367 Coldwater sequences with a 9.6% conflict rate.

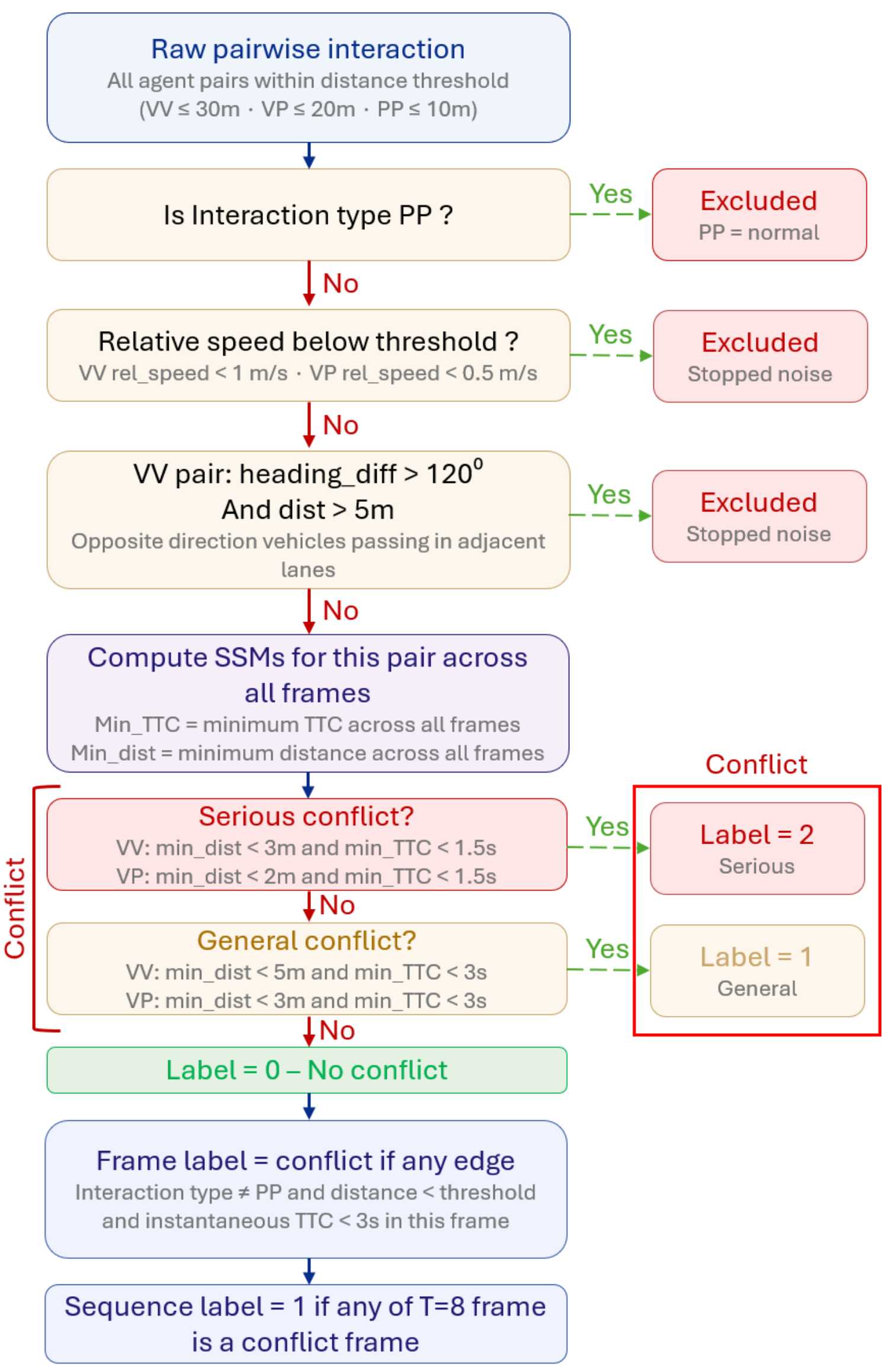

Figure 4: Conflict labeling pipeline from raw pairwise interactions to sequence-level.

Heterogeneous Scene Graph Construction

For each frame $t$, a heterogeneous scene graph $\mathcal{G}_t = (\mathcal{V}_t, \mathcal{E}_t)$ was constructed from the observed road users and qualifying pairwise interactions. Each graph represents the spatial and kinematic state of the intersection and forms the input to the graph encoder.

Each road user was represented as a node $v_i \in \mathcal{V}_t$ with feature vector

$$\mathbf{h}_i^t = \left[x_i^t, y_i^t, v_{x,i}^t, v_{y,i}^t, s_i^t, a_{x,i}^t, a_{y,i}^t, a_i^t, \theta_i^t\right]^\top \tag{97}$$

These features describe local position, velocity, speed, acceleration, and heading, as summarized in Table 5. Agent type was encoded separately as a categorical label identifying vehicle or pedestrian nodes and was used to define heterogeneous relation types.

Each qualifying ordered pair $(i, j)$ was represented by a directed edge $e_{ij}^t \in \mathcal{E}_t$, with source node $i$ and target node $j$. The six-dimensional edge feature vector was defined as:

$$\mathbf{e}_{ij}^t = \left[d_{ij}^t, v_{\mathrm{rel},ij}^t, \mathrm{DRAC}_{ij}^t, \Delta\theta_{ij}^t, \Delta v_{x,ij}^t, \Delta v_{y,ij}^t\right]^\top \tag{98}$$

where

$$v_{x,ij}^t = v_{x,j}^t - v_{x,i}^t, \qquad \Delta v_{y,ij}^t = v_{y,j}^t - v_{y,i}^t \tag{99}$$

Table 6 summarizes the edge attributes. Each edge was assigned one of three relation types: vehicle-vehicle, vehicle-pedestrian, or pedestrian-pedestrian. These labels governed relation-specific message passing and attention.

TTC was excluded from the edge representation because it was used directly in conflict labeling and could therefore introduce target leakage. The retained features describe proximity, relative motion, directionality, and braking demand without explicitly reproducing the TTC criterion. Nevertheless, distance and relative velocity remain components of the labeling process, so the model should be interpreted as learning from the kinematic basis of conflict formation rather than from a fully label-independent feature set. Undefined DRAC values were replaced with 0.0 before normalization to preserve a fixed six-dimensional edge representation.

Table 5: Continuous node features used in each scene graph.

| Feature | Unit | Description |
| --- | --- | --- |
| $(x_i, y_i)$ | m | Local metric coordinates of agent (i), measured eastward and northward relative to the centroid of the region of interest. |
| $(v_{x_i}, v_{y_i})$ | m/s | Longitudinal and lateral velocity components obtained from the derivatives of the Savitzky–Golay-smoothed trajectories. |
| $(\mathrm{speed}_i)$ | m/s | Scalar speed of agent (i), computed as $\left(\sqrt{v_{x_i}^2 + v_{y_i}^2}\right)$. |
| $(a_{x_i}, a_{y_i})$ | m/s² | Acceleration components derived from the temporal derivatives of the smoothed velocity components. |
| $\mathrm{accel}_i$ | m/s² | Magnitude of the acceleration vector, computed as $\left(\sqrt{a_{x_i}^2 + a_{y_i}^2}\right)$. |
| $\mathrm{heading}_i$ | degrees | Direction of motion of agent (i), computed as $\left(\mathrm{atan2}(v_{y_i}, v_{x_i})\right)$ and expressed in degrees. |

Table 6: Continuous edge features used in each scene graph.

| Feature | Unit | Description |
|---|---|---|
| $\text{distance}_{ij}$ | m | Euclidean distance between agents i and j in the projected ground-plane coordinate system. |
| $\text{rel_speed}_{ij}$ | m/s | Magnitude of the relative velocity between the two agents, computed as $\left\|\mathbf{v}_j - \mathbf{v}_i\right\|_2$. |
| $\text{DRAC}_{ij}$ | m/s² | Deceleration rate required to avoid a collision under the prevailing relative-motion conditions. The value was set to zero when DRAC was undefined. |
| $\text{heading_diff}_{ij}$ | degrees | Absolute angular difference between the travel headings of agents i and j. |
| rel $v_{x,ij}$ | m/s | Relative velocity component along the x-axis, calculated as $v_{x,j} - v_{x,i}$. |
| rel $v_{y,ij}$ | m/s | Relative velocity component along the y-axis, calculated as $v_{y,j} - v_{y,i}$. |

Continuous node and edge features were standardized using statistics estimated exclusively from the training partition. For feature $k$,

$$\tilde{x}_k = \frac{x_k - \mu_k^{\text{train}}}{\max\left(\sigma_k^{\text{train}}, \epsilon\right)} \tag{100}$$

where $\mu_k^{\text{train}}$ and $\sigma_k^{\text{train}}$ denote the training-set mean and standard deviation, and $\epsilon$ prevents division by zero for near-constant features. The same parameters were applied to the validation and test partitions without re-estimation to prevent information leakage. Residual missing values were replaced with zero after standardization, corresponding to the training-set mean in the normalized feature space.

Sequence Construction and Data Split

Frame-level scene graphs were arranged into fixed-length sequences using a chronological sliding window applied separately within each video segment. Each sequence contained $T = 8$ consecutive graphs, and the window advanced by four frames. All graphs within a sequence were required to originate from the same video segment, and consecutive graph indices could differ by no more than five frames to preserve temporal continuity.

Sequence labels followed the any-frame rule:

$$y_{\text{seq}} = \max_{t \in \{1,\ldots,T\}} y_t \tag{101}$$

where $y_t \in \{0,1\}$ is the frame-level conflict label. Thus, a sequence was labeled as a conflict if at least one constituent frame was positive. Because adjacent windows overlapped, temporally persistent conflicts could appear in multiple sequences.

This procedure generated 109,028 San Marcos sequences with an overall conflict rate of 18.9%. For Coldwater, 266,901 valid frame graphs produced 47,367 sequences, including 4,536 conflict and 42,831 safe sequences, corresponding to a conflict rate of 9.6%.

Sequences were partitioned at the video-segment level to prevent leakage caused by overlapping windows or temporally adjacent observations from the same event. San Marcos video segments were ordered chronologically and assigned to training, validation, and test partitions according to cumulative sequence counts. The resulting partitions contained 64,945 training, 24,082 validation, and 20,001 test sequences, corresponding to 59.6%, 22.1%, and 18.3% of the dataset, respectively. Their conflict rates were 19.7%, 20.9%, and 13.7%. The actual proportions differed from the nominal targets because complete video segments were retained as indivisible units. The test partition comprised temporally distinct evening recordings. Table 7 summarizes the San Marcos and Coldwater split statistics.

Table 7: Sequence and data-split summary for the San Marcos and Coldwater datasets.

| Site | Partition | Sequences | Proportion | Conflict rate |
|---|---|---|---|---|
| San Marcos | Training | 64,945 | 59.6% | 19.7% |
| San Marcos | Validation | 24,082 | 22.1% | 20.9% |
| San Marcos | Test | 20,001 | 18.3% | 13.7% |
| San Marcos | Total | 109,028 | 100% | 18.9% |
| Coldwater | Training | 32,452 | 68.5 | 10.3% |
| Coldwater | Validation | 9,686 | 20.4 | 9.2% |
| Coldwater | Test | 5,229 | 11.0 | 5.8% |
| Coldwater | Total | 47,367 | 100.0% | 9.6% |

The difference in conflict prevalence between San Marcos and Coldwater provides a useful basis for evaluating whether the learned HERMES representations generalize across sites with different traffic conditions, road-user compositions, and interaction frequencies. Although all three partitions are dominated by safe sequences, the test set contains a smaller proportion of conflict sequences than the training and validation subsets. This distribution reinforces the need to evaluate model performance using class-imbalance-sensitive and false-alarm-aware metrics rather than relying on accuracy alone.

## 5 RESULTS

### 5.1 Overall Comparative Performance

The benchmark compared 11 models across three model families: conventional machine-learning models, temporal deep-learning models, and graph-based architectures. The machine-learning models operated on sequence-level descriptors, the deep-learning models processed fixed-length temporal feature sequences, and the graph models explicitly represented heterogeneous road-user interactions. Table 8 reports the mean test performance and corresponding standard deviation across five random seeds. Figure 5 and Figure 6 present the five-seed aggregated ROC and precision-recall curves for representative models from each model family.

Machine-Learning Tier

Among the five machine-learning models, XGBoost achieved the highest AUC-ROC at $0.9569 \pm 0.0005$, followed closely by LightGBM at $0.9565 \pm 0.0004$ and RF at $0.9559 \pm 0.0001$. XGBoost also produced the highest AUC-PR and F1 score within this tier, reaching $0.7270 \pm 0.0045$ and $0.7085 \pm 0.0029$, respectively. Logistic Regression and SVM yielded lower AUC-ROC values of $0.9341 \pm 0.0002$ and $0.9363 \pm 0.0009$, indicating that their decision functions were less effective at representing the nonlinear conflict patterns contained in the handcrafted feature space.

The constrained false-alarm results revealed greater operational differences than the AUC-ROC values alone. At a 5% false-alarm rate, RF achieved the highest sensitivity among the machine-learning models at $0.6197 \pm 0.0057$, followed by LightGBM at $0.6171 \pm 0.0028$ and XGBoost at $0.6152 \pm 0.0078$. At a 10% false-alarm rate, RF again produced the highest sensitivity within this tier at $0.9077 \pm 0.0039$, compared with $0.9055 \pm 0.0041$ for XGBoost and $0.9023 \pm 0.0041$ for LightGBM. Thus, although the ensemble models achieved similar overall discrimination, their sensitivities remained limited under the more stringent 5% false-alarm constraint.

Deep-Learning Tier

All four temporal deep-learning models improved upon the strongest machine-learning results in AUC-ROC, AUC-PR, and low-false-alarm sensitivity. AUC-ROC ranged from$0.9677 \pm 0.0024$ for the TCN to $0.9743 \pm 0.0028$ for the Transformer. The Transformer also achieved the highest AUC-PR within the deep-learning tier at $0.8375 \pm 0.0248$, followed by the GRU at $0.8116 \pm 0.0093$, the BiLSTM at $0.8023 \pm 0.0037$, and the TCN at $0.7991 \pm 0.0204$.

At a $5\%$ false-alarm rate, the Transformer detected $80.96\%$ of conflict sequences, compared with $61.97\%$ for RF, which was the strongest machine-learning model under this criterion. The GRU and BiLSTM achieved sensitivities of $0.7687 \pm 0.0088$ and $0.7629 \pm 0.0054$, respectively, while the TCN reached $0.7619 \pm 0.0243$. These results indicate that modeling temporal evolution across the eight-frame sequence provided predictive information beyond the aggregated descriptors used by the conventional machine-learning models.

Variability differed across the temporal architectures. The Transformer exhibited the largest standard deviation in AUC-PR at $0.0248$, whereas the TCN showed the greatest variability in S@5\%FAR at $0.0243$ and comparatively high variability across several other metrics. The BiLSTM and GRU produced more consistent results across the five seeds, with generally smaller standard deviations.

Graph Tier

Baseline HERMES achieved an AUC-ROC of $0.9692 \pm 0.0011$, compared with $0.9743 \pm 0.0028$ for the Transformer. Despite its slightly lower overall AUC-ROC, Baseline HERMES produced higher AUC-PR, F1, and sensitivity at the more restrictive $5\%$ false-alarm operating point. Specifically, Baseline HERMES achieved an AUC-PR of $0.8436 \pm 0.0038$, an F1 score of $0.7585 \pm 0.0060$, and an S@5\%FAR of $0.8160 \pm 0.0058$, compared with $0.8375 \pm 0.0248$, $0.7540 \pm 0.0075$, and $0.8096 \pm 0.0186$, respectively, for the Transformer.

The relative performance of the two models changed when the allowable false-alarm rate increased. At $10\%$ FAR, the Transformer achieved a sensitivity of $0.9606 \pm 0.0064$, whereas Baseline HERMES reached $0.9263 \pm 0.0076$. This reversal indicates that neither model uniformly dominated the other throughout the low-false-positive-rate region. Baseline HERMES provided a modest advantage when false alarms were constrained more tightly, whereas the Transformer recovered additional conflict sequences more rapidly as the operating threshold was relaxed. From an operational perspective, the preferred model therefore depends on the acceptable false-alarm tolerance: Baseline HERMES may be more suitable under stringent false-alarm control, while the Transformer may be preferable when a higher false-alarm rate can be tolerated to obtain greater sensitivity.

Although these results suggest that explicit road-user interaction modeling may benefit minority conflict retrieval under stringent operating conditions, the comparison does not isolate the contribution of graph structure because the two models differ in architecture, parameterization, and input representation.

Table 8: Comparative multi-seed benchmark performance across all model tiers.

| Model | AUC-ROC | AUC-PR | S@5%FAR | S@10%FAR | F1 | MCC | Kappa | Brier score |
|---|---|---|---|---|---|---|---|---|
| **Logistic Regression** | 0.9341 ± 0.0002 | 0.5815 ± 0.0003 | 0.4726 ± 0.0002 | 0.7713 ± 0.0004 | 0.6472 ± 0.0002 | 0.6124 ± 0.0033 | 0.5709 ± 0.0026 | 0.1023 ± 0.0002 |
| **Random Forest** | 0.9559 ± 0.0001 | 0.7080 ± 0.0013 | 0.6197 ± 0.0057 | 0.9077 ± 0.0039 | 0.6309 ± 0.0023 | 0.5768 ± 0.0024 | 0.5758 ± 0.0025 | 0.0581 ± 0.0015 |
| **XGBoost** | 0.9569 ± 0.0005 | 0.7270 ± 0.0045 | 0.6152 ± 0.0078 | 0.9055 ± 0.0041 | 0.7085 ± 0.0029 | 0.6663 ± 0.0036 | 0.6529 ± 0.0033 | 0.0668 ± 0.0004 |
| **LightGBM** | 0.9565 ± 0.0004 | 0.7240 ± 0.0022 | 0.6171 ± 0.0028 | 0.9023 ± 0.0033 | 0.6998 ± 0.0022 | 0.6535 ± 0.0025 | 0.6439 ± 0.0008 | 0.0698 ± 0.00022 |
| **SVM** | 0.9363 ± 0.0009 | 0.6226 ± 0.0012 | 0.5284 ± 0.0040 | 0.8093 ± 0.0038 | 0.6109 ± 0.0003 | 0.5485 ± 0.0004 | 0.5484 ± 0.0004 | 0.0706 ± 0.0018 |
| **BiLSTM** | 0.9688 ± 0.0005 | 0.8023 ± 0.0037 | 0.7629 ± 0.0054 | 0.9502 ± 0.0025 | 0.7469 ± 0.0021 | 0.7129 ± 0.0019 | 0.6988 ± 0.0035 | 0.0614 ± 0.0026 |
| **GRU** | 0.9700 ± 0.0011 | 0.8116 ± 0.0093 | 0.7687 ± 0.0088 | 0.9521 ± 0.0038 | 0.7507 ± 0.0045 | 0.7179 ± 0.0038 | 0.7031 ± 0.0060 | 0.0600 ± 0.0021 |
| **TCN** | 0.9677 ± 0.0024 | 0.7991 ± 0.0204 | 0.7619 ± 0.0243 | 0.9417 ± 0.0111 | 0.7420 ± 0.0107 | 0.7073 ± 0.0125 | 0.6931 ± 0.0132 | 0.0602 ± 0.0024 |
| **Transformer** | 0.9743 ± 0.0028 | 0.8375 ± 0.0248 | 0.8096 ± 0.0186 | 0.9606 ± 0.0064 | 0.7540 ± 0.0075 | 0.7259 ± 0.0078 | 0.7057 ± 0.0094 | 0.0624 ± 0.0024 |
| **Baseline HERMES** | 0.9692 ± 0.0011 | 0.8436 ± 0.0038 | 0.8160 ± 0.0058 | 0.9263 ± 0.0076 | 0.7585 ± 0.0060 | 0.7228 ± 0.0049 | 0.7145 ± 0.0085 | 0.0609 ± 0.0040 |
| **Enhanced HERMES** | **0.9898 ± 0.0013** | **0.9412 ± 0.0067** | **0.9574 ± 0.0075** | **0.9931 ± 0.0022** | **0.8449 ± 0.0103** | **0.8248 ± 0.0108** | **0.8168 ± 0.0126** | **0.0368 ± 0.0031** |

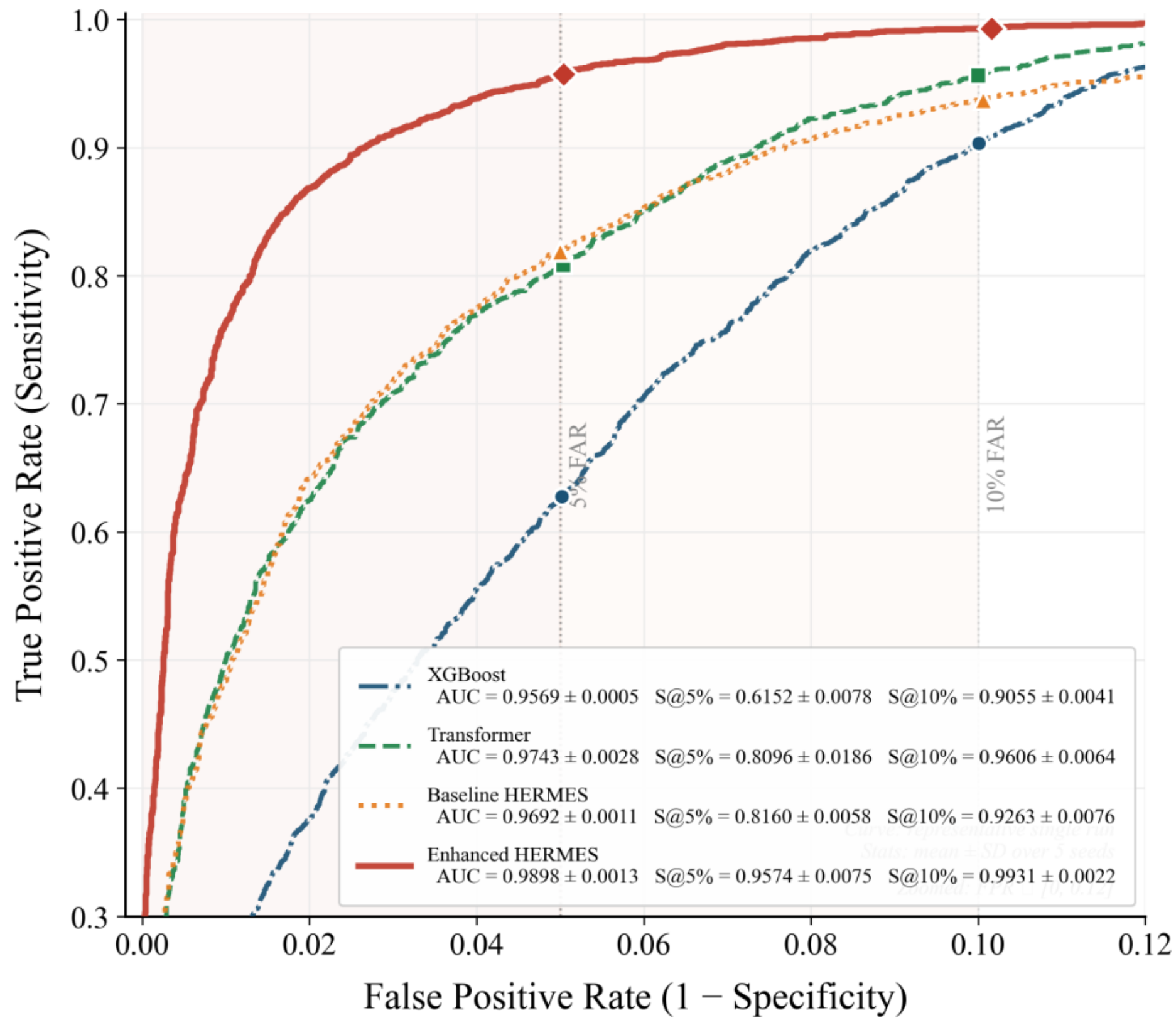

Figure 5: Receiver operating characteristic curves for selected models.

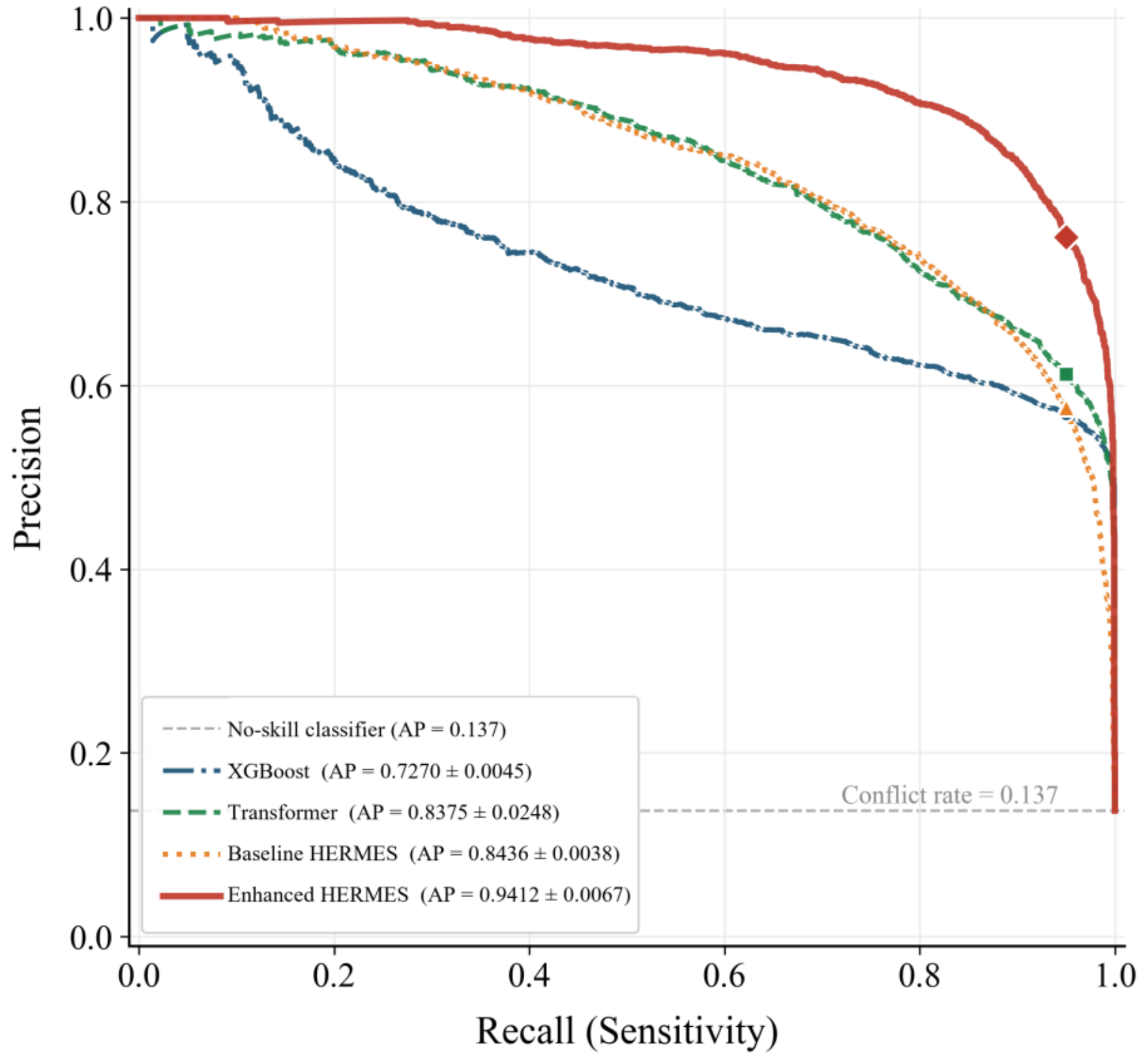

Figure 6: Precision-recall curves for selected models.

Enhanced HERMES

Enhanced HERMES achieved the strongest performance across all eight reported metrics. Its AUC-ROC reached $0.9898 \pm 0.0013$, representing absolute improvements of $0.0329$ over XGBoost, $0.0155$ over the Transformer, and $0.0206$ over Baseline HERMES. The corresponding AUC-PR was $0.9412 \pm 0.0067$, exceeding the strongest machine-learning, deep-learning, and baseline graph results by $0.2142$, $0.1037$, and $0.0976$, respectively.

The largest practical separation was observed under the constrained false-alarm criteria. Enhanced HERMES achieved an $S@5\%FAR$ of $0.9574 \pm 0.0075$, compared with $0.8160 \pm 0.0058$ for Baseline HERMES, $0.8096 \pm 0.0186$ for the Transformer, and $0.6197 \pm 0.0057$ for RF, which was the strongest machine-learning model under this criterion. These values correspond to absolute sensitivity improvements of $0.1414$, $0.1478$, and $0.3377$, respectively. At a $10\%$ false-alarm rate, Enhanced HERMES achieved a sensitivity of $0.9931 \pm 0.0022$, compared with $0.9606 \pm 0.0064$ for the Transformer, $0.9263 \pm 0.0076$ for Baseline HERMES, and $0.9077 \pm 0.0039$ for RF.

Enhanced HERMES also achieved the highest F1 score, MCC, and Cohen's kappa values, reaching $0.8449 \pm 0.0103$, $0.8248 \pm 0.0108$, and $0.8168 \pm 0.0126$, respectively. These results indicate that the model maintained a comparatively strong balance between conflict detection and safe-sequence classification rather than improving sensitivity solely through excessive positive predictions.

The Brier score of $0.0368 \pm 0.0031$ was the lowest among the evaluated models, indicating the smallest mean squared error between the predicted probabilities and observed binary outcomes and, therefore, improved probabilistic accuracy. Enhanced HERMES also exhibited comparatively small standard deviations across the five seeds, particularly for AUC-ROC and AUC-PR, suggesting that its ranking performance remained relatively stable across the evaluated random initializations.

ROC and PR Curve Analysis

The ROC curves show that all four selected models performed substantially above the random-classifier reference. Enhanced HERMES remained closest to the upper-left boundary and exhibited the clearest separation within the low-false-positive-rate region. This pattern is consistent with its five-seed $S@5\%FAR$ of $0.9574 \pm 0.0075$, compared with $0.8160 \pm 0.0058$ for Baseline HERMES, $0.8096 \pm 0.0186$ for the Transformer, and $0.6152 \pm 0.0078$ for XGBoost. The relative positions of Baseline HERMES and the Transformer changed between the $5\%$ and $10\%$ FAR operating points, illustrating that their comparative performance depended on the selected false-alarm constraint.

The PR curves provide a more direct assessment of conflict retrieval under the $13.7\%$ conflict prevalence in the test set. XGBoost exhibited a more pronounced decline in precision as recall increased, whereas the Transformer and Baseline HERMES maintained higher precision across a broader recall range. Enhanced HERMES produced the strongest PR profile, consistent with its five-seed AUC-PR of $0.9412 \pm 0.0067$. The horizontal no-skill reference corresponds to the observed conflict prevalence and therefore represents the expected precision of a classifier without discriminatory ability.

**5.2 Statistical Stability and Multi-Seed Robustness**

Because neural-network performance may vary with weight initialization, mini-batch ordering, and stochastic optimization, the six deep-learning and graph-based models were evaluated across five random seeds. Mean performance and the corresponding standard deviations are reported in Table 8, while Figure 7 shows the seed-level distributions of AUC-ROC.

The evaluated models generally exhibited limited run-to-run variation, although the magnitude differed across architectures. The Transformer and TCN showed the greatest variability in AUC-PR, with standard deviations of 0.0248 and 0.0204, respectively. They also exhibited comparatively high variability in sensitivity at 5% FAR. In contrast, Baseline HERMES produced standard deviations of 0.0011 for AUC-ROC and 0.0038 for AUC-PR. Enhanced HERMES remained similarly stable, with corresponding standard deviations of 0.0013 and 0.0067.

As illustrated in Figure 7, the five Enhanced HERMES runs formed a compact distribution centered at an AUC-ROC of 0.9898, with no overlap with the observed seed-level ranges of the Transformer or Baseline HERMES. Its mean AUC-ROC exceeded those models by 0.0155 and 0.0206, respectively, substantially larger than the corresponding inter-seed standard deviations. Enhanced HERMES also achieved a higher AUC-ROC than each comparator in all five evaluated runs.

These results indicate that the performance advantage of Enhanced HERMES was reproduced across the tested random seeds and was not attributable to a single favorable initialization. Nevertheless, the five-run analysis should be interpreted as evidence of empirical training stability rather than formal statistical significance.

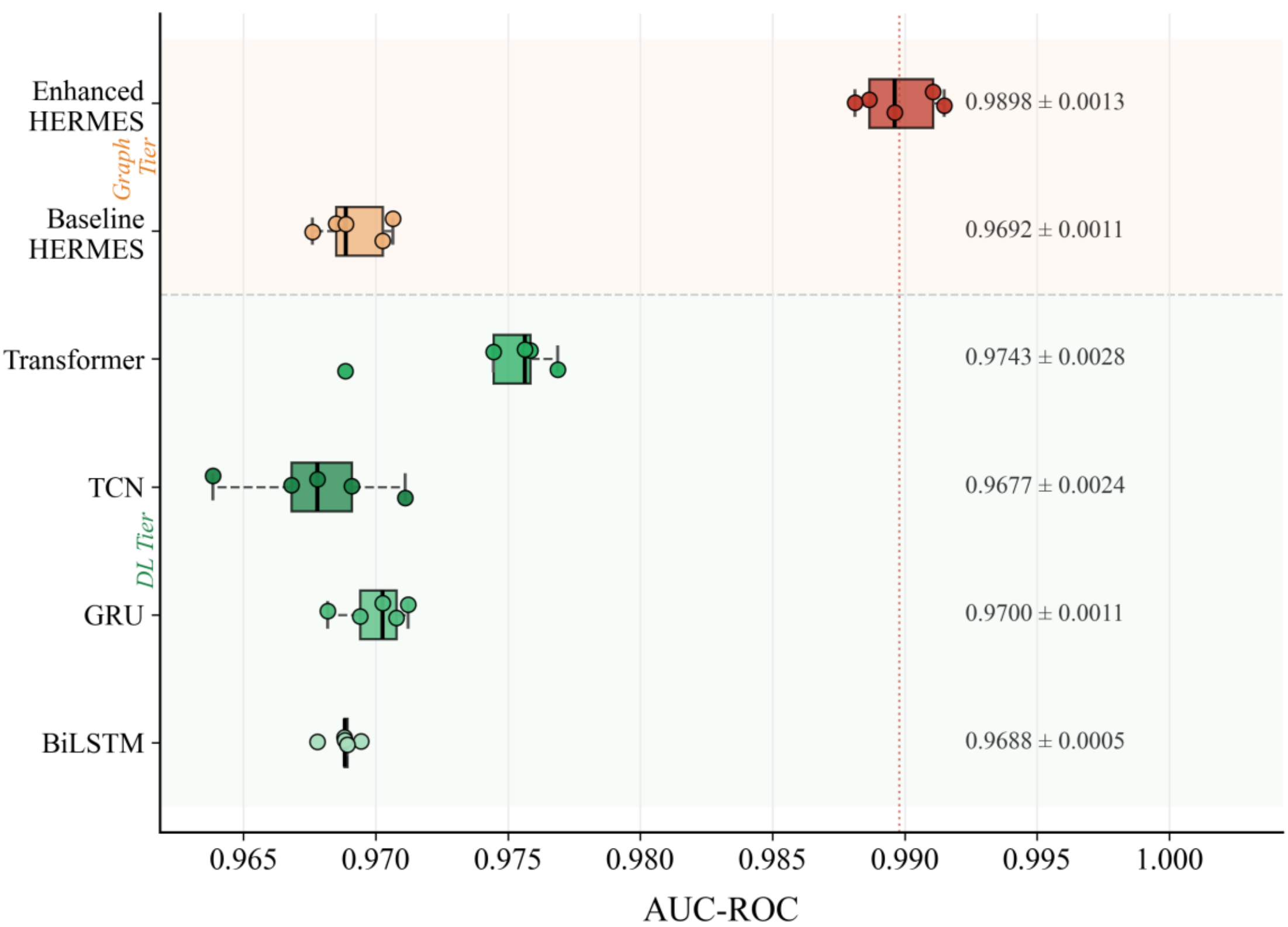

Figure 7: Seed-level AUC-ROC distributions for the stochastic deep-learning and graph-based models across five random seeds.

**5.3 Low-False-Positive Operational Sensitivity**

Global discrimination metrics summarize ranking performance across all decision thresholds, but they do not fully characterize model behavior under restrictive false-alarm conditions. This distinction is important for proactive traffic-conflict monitoring, where frequent false warnings may reduce operational usefulness. Accordingly, low-false-alarm sensitivity was assessed at false-alarm rates of 5% and 10%, as reported in Table 8 and illustrated by the operating points in Figure 5.

At 5% FAR, Enhanced HERMES achieved a sensitivity of $0.9574 \pm 0.0075$, compared with $0.8160 \pm 0.0058$ for Baseline HERMES, $0.8096 \pm 0.0186$ for the Transformer, and $0.6152 \pm 0.0078$ for XGBoost. These values correspond to absolute improvements of 14.14, 14.78, and 34.22 percentage points, respectively. Thus, under a false-alarm allowance equivalent to incorrectly flagging approximately one in twenty safe sequences, Enhanced HERMES retained substantially higher conflict sensitivity than the comparison models.

The advantage persisted at 10% FAR. Enhanced HERMES reached a sensitivity of $0.9931 \pm 0.0022$, exceeding the Transformer at $0.9606 \pm 0.0064$, Baseline HERMES at $0.9263 \pm 0.0076$, and XGBoost at $0.9055 \pm 0.0041$. The ROC profiles further show that the proposed model attained high true-positive rates earlier in the low-FPR region, indicating a more favorable trade-off between conflict retrieval and false alarms.

These results demonstrate that the performance advantage of Enhanced HERMES was not limited to global ranking metrics. The model also maintained high conflict-detection sensitivity under restrictive false-alarm constraints, which is particularly relevant for warning and monitoring applications in which excessive false positives may limit practical deployment.

### 5.4 Confusion Matrix and Error Trade-off Analysis

ROC and PR analyses summarize model performance across decision thresholds, whereas confusion matrices show the classification errors produced at a specific operating threshold. Figure 8 presents the confusion matrices for XGBoost, the Transformer, Baseline HERMES, and Enhanced HERMES on the held-out test set.

XGBoost correctly classified 2,323 of the 2,742 conflict sequences, corresponding to a conflict recall of 84.7%, but incorrectly flagged 1,491 safe sequences. The Transformer increased conflict recalls to 94.8%, reducing false negatives from 419 to 142, although its false-positive count increased to 1,645. Baseline HERMES produced fewer false positives than both models, with 1,224 safe sequences incorrectly classified, while retaining a conflict recall of 88.5%.

Enhanced HERMES provided the most balanced error profile. It correctly classified 95.5% of safe sequences and 94.7% of conflict sequences, producing 769 false positives and 146 false negatives. Relative to XGBoost, this represents reductions of 48.4% in false positives and 65.2% in false negatives. Compared with the Transformer, Enhanced HERMES reduced false positives by 53.3%, while conflict recall differed by only 0.1 percentage points (94.7% versus 94.8%). Compared with Baseline HERMES, it reduced both false positives and false negatives, from 1,224 to 769 and from 316 to 146, respectively.

These results indicate that Enhanced HERMES achieved the strongest combined control of missed conflicts and false alarms at the evaluated threshold. The improvement complements the threshold-independent and low-false-alarm results by showing that the model's higher discrimination translated into a more favorable classification error profile at the selected operating point.

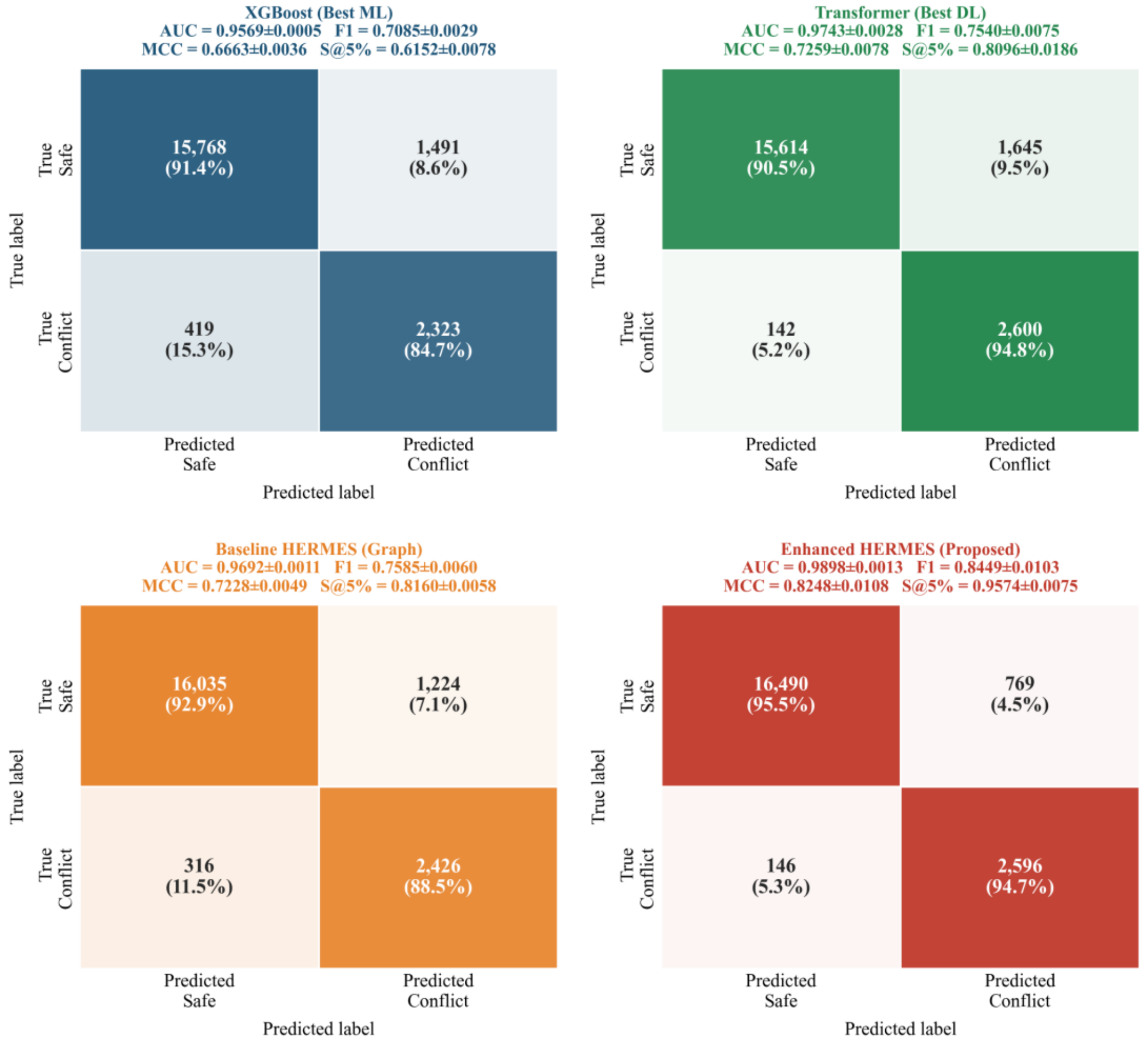

Figure 8: Comparative confusion matrices of the four strongest benchmark models.

### 5.5 Ablation Analysis of HERMES Components

A structured ablation study was conducted to assess four design dimensions of HERMES: edge-feature composition, graph architecture, temporal encoding, and interaction-type coverage. To ensure a consistent and computationally tractable comparison across the full ablation grid, each variant was trained using the same fixed 10-epoch budget over five random seeds. For each seed, the checkpoint achieving the highest validation AUC-ROC within this training horizon was retained and evaluated on the unchanged held-out test set. This controlled protocol ensured that all variants were compared under identical optimization conditions, allowing the relative contribution of each component to be assessed without confounding differences in training effort. Table 9 reports the quantitative results of the multi-seed ablation study.

Table 9: Multi-seed ablation performance summary of HERMES design components

| Group | Variant | Description | AUC-ROC | AUC-PR | S@5%FAR |
|---|---|---|---|---|---|
| **SSM Features** | A0 | No informative edge SSM features | 0.9541 ± 0.0013 | 0.7761 ± 0.0088 | 0.7326 ± 0.0051 |
| | A1 | Relational kinematics only | 0.9646 ± 0.0028 | 0.8068 ± 0.0144 | 0.7683 ± 0.0158 |
| | A2 | Kinematics + DRAC | 0.9611 ± 0.0011 | 0.7917 ± 0.0070 | 0.7492 ± 0.0100 |
| | A3 | Kinematics + DRAC + heading | 0.9616 ± 0.0022 | 0.8038 ± 0.0040 | 0.7527 ± 0.0059 |
| | A4 | Full edge feature set | 0.9616 ± 0.0022 | 0.7993 ± 0.0140 | 0.7511 ± 0.0211 |
| **Architecture** | B1 | No graph relational module | 0.8888 ± 0.0011 | 0.5230 ± 0.0055 | 0.4222 ± 0.0115 |
| | B2 | Simple graph convolution | 0.9319 ± 0.0039 | 0.6435 ± 0.0151 | 0.5544 ± 0.0268 |
| | B3 | SSM-guided graph attention + mean pooling | 0.9675 ± 0.0020 | 0.8158 ± 0.0077 | 0.7884 ± 0.0230 |
| | B4 | Full HERMES graph architecture | 0.9616 ± 0.0022 | 0.7993 ± 0.0140 | 0.7511 ± 0.0211 |
| **Temporal Encoder** | C1 | Single frame (no temporal modeling) | 0.8516 ± 0.0023 | 0.5877 ± 0.0146 | 0.5533 ± 0.0104 |
| | C2 | GNN + GRU | 0.9617 ± 0.0013 | 0.7969 ± 0.0043 | 0.7479 ± 0.0118 |
| | C3 | GNN + Transformer | 0.9614 ± 0.0022 | 0.7994 ± 0.0120 | 0.7425 ± 0.0170 |
| | C4 | GNN + LSTM (final) | 0.9616 ± 0.0022 | 0.7993 ± 0.0140 | 0.7511 ± 0.0211 |
| **Interaction Types** | D1 | Vehicle–vehicle only | 0.9614 ± 0.0022 | 0.7993 ± 0.0115 | 0.7587 ± 0.0115 |
| | D2 | Vehicle–pedestrian only | 0.9144 ± 0.0048 | 0.5897 ± 0.0107 | 0.4877 ± 0.0198 |
| | D3 | Vehicle–vehicle + vehicle–pedestrian | 0.9617 ± 0.0025 | 0.8069 ± 0.0141 | 0.7498 ± 0.0243 |
| | D4 | All relation types (final) | 0.9616 ± 0.0022 | 0.7993 ± 0.0140 | 0.7511 ± 0.0211 |

Edge-Feature Composition

Variant A0 removed graph connectivity by setting the relation-specific adjacency matrices to zero. The model therefore relied only on node-level trajectory states, while the dummy edge channel was retained for architectural compatibility. This configuration achieved an AUC-ROC of $0.9541 \pm 0.0013$. Introducing pairwise distance and relative speed in A1 increased AUC-ROC to $0.9646 \pm 0.0028$, corresponding to an absolute improvement of $0.0105$. A1 also achieved the highest AUC-PR and $S@5\%FAR$ within this group.

The results show that explicit relational kinematics provide substantial information beyond isolated agent states. DRAC, heading difference, and relative-velocity components preserve additional safety-relevant information on braking demand, directional alignment, and motion orientation. Although these descriptors did not yield further gains under the reduced ablation protocol, their inclusion supports the complete physical representation of heterogeneous interactions adopted in HERMES.

Graph Architecture

The non-graph configuration achieved an AUC-ROC of $0.8888 \pm 0.0011$, indicating a substantial performance reduction when relational interaction modeling was removed. Introducing a simple graph convolution increased AUC-ROC to $0.9319 \pm 0.0039$. The SSM-guided attention configuration with mean pooling further improved performance to $0.9675 \pm 0.0020$, representing an absolute gain of $0.0356$ over the simple graph model.

The full HERMES architecture achieved an AUC-ROC of $0.9616 \pm 0.0022$ and retained the principal gains introduced by relational message passing and SSM-guided attention. These results identify interaction-aware graph processing and safety-informed attention weighting as the most influential architectural elements. The complete architecture integrates these components with dynamic edge refinement and safety-aware pooling to form the final relational representation used by HERMES.

Temporal Encoder

The single-frame configuration produced an AUC-ROC of $0.8516 \pm 0.0023$, demonstrating that a static scene snapshot does not capture the short-term motion evolution required for reliable conflict discrimination. Introducing temporal encoding increased AUC-ROC to approximately $0.962$ for all three sequence models.

The GRU achieved the highest AUC-ROC at $0.9617 \pm 0.0013$, the Transformer achieved the highest AUC-PR at $0.7994 \pm 0.0120$, and the LSTM produced the highest $S@5\%FAR$ at $0.7511 \pm 0.0211$. The small differences among the temporal variants indicate that the inclusion of temporal memory is more influential than the specific encoder type. HERMES adopts the LSTM to capture short-term interaction evolution within the integrated graph-temporal architecture.

Interaction-Type Coverage

The VV-only configuration achieved an AUC-ROC of $0.9614 \pm 0.0022$ and the highest $S@5\%FAR$ in this group at $0.7587 \pm 0.0115$. The VP-only configuration reached an AUC-ROC of $0.9144 \pm 0.0048$, whereas combining VV and VP relations produced the highest AUC-ROC and AUC-PR values, at $0.9617 \pm 0.0025$ and $0.8069 \pm 0.0141$, respectively. Including PP relations resulted in a comparable AUC-ROC of $0.9616 \pm 0.0022$.

These findings indicate that VV relations provide a strong contextual representation of scene dynamics, while vehicle-pedestrian relations contribute complementary information associated with direct pedestrian exposure. The full relation configuration retains all three interaction types, enabling the model to represent both conflict-relevant and contextual interactions within a unified heterogeneous graph.

Overall, the ablation results demonstrate that HERMES derives its performance primarily from the coordinated integration of relational graph modeling, SSM-guided attention, and temporal sequence encoding. Removing graph interactions or temporal information produced the largest performance losses, whereas introducing pairwise kinematics and attention-based message weighting yielded substantial improvements. The complete HERMES configuration combines these high-impact components with the full safety-informed edge representation and heterogeneous interaction structure, providing a unified framework for modeling the spatial, relational, and temporal characteristics of traffic conflicts.

**5.6 Cross-Site Transferability Analysis**

The cross-site transferability of Enhanced HERMES was evaluated using San Marcos as the source site and Coldwater as the external target site. The Coldwater dataset contained 47,367 sequences with an overall conflict rate of 9.6%. Its fixed validation and test partitions contained 9,686 and 5,229 sequences, with conflict rates of 9.2% and 5.8%, respectively.

Two complementary experiments were conducted. First, zero-shot transfer was assessed by applying the San Marcos-trained Enhanced HERMES model directly to the Coldwater test set without using any Coldwater training data. Second, mixed-source training was performed by combining all 64,945 San Marcos training sequences with sequences drawn from randomly selected subsets of the 93 Coldwater training video segments. The selected Coldwater proportions ranged from 10% to 60% of the available training segments. For each ratio, a new Enhanced HERMES model was trained from random initialization across three seeds. The checkpoint with the highest AUC-ROC on the fixed Coldwater validation set was retained and subsequently evaluated on the unchanged Coldwater test set.

Because the sampling ratios were defined at the video-segment level, the number of Coldwater sequences included at a given ratio varied across random seeds. Thus, the experiments represent progressive inclusion of target-site segments rather than fixed proportions of target-site sequences.

Zero-Shot Transfer

Without exposure to Coldwater training data, Enhanced HERMES achieved an AUC-ROC of 0.9752, an AUC-PR of 0.7829, an F1 score of 0.7188, and an $S@5\%FAR$ of 0.9145. The corresponding Brier score was 0.0300. At the evaluated classification threshold, the model achieved a precision of 0.6620, a recall of 0.7862, and a specificity of 0.9752.

These results indicate that the interaction and temporal representations learned from San Marcos retained substantial predictive value when transferred directly to a geographically distinct site with a lower conflict prevalence. The lower AUC-PR and F1 values relative to the source-site benchmark suggest that cross-site differences affected minority conflict identification more strongly than overall ranking performance.

Mixed-Source Training

Table 10 summarizes the performance obtained when all San Marcos training data were combined with progressively larger subsets of the Coldwater training segments.

Table 10: Coldwater test performance under zero-shot transfer and mixed-source training.

| **Training segments** | **AUC-ROC** | **AUC-PR** | **F1** | **S@5%FAR** |
|---|---|---|---|---|
| Zero-shot | 0.9752 | 0.7829 | 0.7188 | 0.9145 |
| 10% segments | 0.9901 ± 0.0034 | 0.8750 ± 0.0313 | 0.7977 ± 0.0280 | 0.9594 ± 0.0346 |
| 20% segments | 0.9903 ± 0.0035 | 0.8389 ± 0.0619 | 0.7861 ± 0.0424 | 0.9869 ± 0.0057 |
| 30% segments | 0.9885 ± 0.0061 | 0.8415 ± 0.1049 | 0.7719 ± 0.0653 | 0.9539 ± 0.0461 |
| 40% segments | 0.9879 ± 0.0012 | 0.8597 ± 0.0243 | 0.7564 ± 0.0363 | 0.9463 ± 0.0187 |
| 50% segments | 0.9875 ± 0.0080 | 0.8753 ± 0.0586 | 0.7925 ± 0.0562 | 0.9726 ± 0.0083 |
| 60% segments | 0.9920 ± 0.0022 | 0.8857 ± 0.0422 | 0.7831 ± 0.0499 | 0.9803 ± 0.0057 |

Mixed-source training improved performance relative to zero-shot transfer at every evaluated Coldwater inclusion ratio. The highest mean AUC-ROC and AUC-PR were obtained using 60% of the Coldwater training segments, reaching $0.9920 \pm 0.0022$ and $0.8857 \pm 0.0422$, respectively. The 10% setting achieved the highest mean F1 score at $0.7977 \pm 0.0280$, while the 20% setting produced the highest $S@5\%FAR$ at $0.9869 \pm 0.0057$.

Performance did not increase monotonically with the nominal Coldwater ratio. This variation is expected because the ratios represent proportions of video segments rather than equal numbers of sequences. The sampled subsets differed in duration, conflict prevalence, and interaction composition across seeds. Accordingly, transfer performance depended on both the quantity and representativeness of the selected target-site segments.

Overall, the results demonstrate that Enhanced HERMES retained strong predictive performance under direct zero-shot transfer and achieved additional gains when source-site training data were jointly modeled with limited target-site information. The improvement observed with only 10% of the Coldwater training segments indicates that relatively modest target-site coverage can substantially strengthen cross-site conflict prediction.

## 6 CONCLUSION

This study developed HERMES, a heterogeneous edge-relational graph-learning framework for scene-level traffic conflict classification at signalized intersections. The framework represents vehicles and pedestrians as heterogeneous nodes and models vehicle-vehicle, vehicle-pedestrian, and pedestrian-pedestrian interactions through relation-specific edges containing continuous kinematic and surrogate-safety descriptors. By integrating relational attention, node-edge representation learning, graph-level aggregation, and temporal sequence encoding, HERMES preserves the spatial, relational, and short-term temporal characteristics of multi-agent traffic scenes.

Enhanced HERMES achieved the highest overall performance among the evaluated machine-learning, temporal deep-learning, and graph-based models on the held-out San Marcos dataset. It obtained an AUC-ROC of 0.9898±0.0013, an AUC-PR of 0.9412±0.0067, an F1 score of 0.8449±0.0103, and a Brier score of 0.0368±0.0031. These results indicate that preserving agent identities and interaction topology provides information that is not fully retained by flat feature representations. The ablation analysis further showed that relational graph processing, SSM-informed attention, and temporal sequence encoding were the principal contributors to conflict discrimination. Distance and relative speed supplied much of the immediately discriminative interaction information, while the complete edge representation retained a broader description of interaction geometry, directional variation, and braking demand.

The independent-site evaluation demonstrated that the learned representation remained effective when transferred from San Marcos to the broadly comparable Coldwater intersection. Without target-site training, Enhanced HERMES achieved an AUC-ROC of 0.9752 and an AUC-PR of 0.7829. Joint training with selected Coldwater segments produced additional improvements, although performance varied with the quantity and interaction composition of the included target-site data. These findings suggest that limited representative target-site observations can support adaptation while retaining the relational information learned from the source location.

Overall, the study demonstrates the value of combining heterogeneous scene graphs, safety-relevant interaction attributes, and temporal modeling for trajectory-based conflict classification. The framework provides a structured basis for continuous intersection-level safety assessment and for identifying evolving multi-agent interaction patterns from roadside video. Because the present outcomes were constructed from trajectory-derived SSM criteria and evaluated at two comparable intersections, future research should examine expert-validated conflict outcomes, additional roadway geometries and traffic conditions, longer observation horizons, event-level alert performance, and real-time computational requirements. These extensions will support further assessment of HERMES as a transferable component of proactive roadside traffic-safety monitoring systems.

## AUTHOR CONTRIBUTION STATEMENT

Md Monzurul Islam: Writing – original draft, Visualization, Validation, Software, Methodology, Formal analysis, Data curation, Conceptualization. Subasish Das: Writing – review & editing, Validation, Supervision, Resources, Project administration. All authors reviewed the results and approved the final version of the manuscript.

## DECLARATION OF GENERATIVE AI AND AI-ASSISTED TECHNOLOGIES IN THE WRITING PROCESS

Generative AI was not used in a substantial manner for writing the manuscript.

## DECLARATION OF COMPETING INTEREST

The authors declare that they have no known competing financial interests or personal relationships that could have appeared to influence the work reported in this paper.

## DATA AND CODE AVAILABILITY

The source code is publicly available at https://github.com/Xatta-Trone/HERMES. The full research dataset is not publicly available; however, it may be made available from the corresponding author upon reasonable request.

## FUNDING SOURCES

This research did not receive any specific grant from funding agencies in the public, commercial, or non-profit sectors.